\pdfoutput=1
\documentclass[11pt]{article}

\usepackage[final]{acl}
\usepackage{xcolor}
\usepackage{times}
\usepackage{latexsym}
\usepackage[most]{tcolorbox}
\definecolor{maincolor}{HTML}{00224e} 
\usepackage[T1]{fontenc}
\usepackage[utf8]{inputenc}

\usepackage{microtype}

\usepackage{inconsolata}

\usepackage{graphicx}

\usepackage{array}
\usepackage{longtable}
\usepackage{booktabs}
\usepackage{multirow}
\usepackage{graphicx} 
\usepackage{siunitx}  

\usepackage{multicol}

\title{Unraveling SITT: Social Influence Technique Taxonomy \\ and Detection with LLMs}

\author{Wiktoria Mieleszczenko-Kowszewicz \\
  Warsaw University of Technology / Address line 1 \\
  \texttt{email@domain} \\\And
Beata Bajcar \\
  Wrocław University of Science and Technology / Address line 1 \\
  \texttt{email@domain} \\\And
Aleksander Szczęsny \\
  Wrocław University of Science and Technology / Address line 1 \\
  \texttt{email@domain} \\\And
Maciej Markiewicz\\
  Wrocław University of Science and Technology / Address line 1 \\
  \texttt{email@domain} \\\And
Jolanta Babiak\\
  Wrocław University of Science and Technology / Address line 1 \\
  \texttt{email@domain} \\\And
Berenika Dyczek\\
  University of
Wrocław / Address line 1 \\
  Lincoln University College, Petaling Jaya, Malaysian/ Address line 2 \\
  \texttt{email@domain} \\\And
Przemysław Kazienko\\
  Wrocław University of Science and Technology / Address line 1 \\
  \texttt{email@domain} \\\And}
  
\author{
 \textbf{Wiktoria Mieleszczenko-Kowszewicz\textsuperscript{1}},
 \textbf{Beata Bajcar\textsuperscript{2}},
 \textbf{Aleksander Szczęsny\textsuperscript{2}}, \\
 \textbf{Maciej Markiewicz\textsuperscript{2}}, 
 \textbf{Jolanta Babiak\textsuperscript{2}},
 \textbf{Berenika Dyczek\textsuperscript{3}},
 \textbf{Przemysław Kazienko\textsuperscript{2}},
\\
 \textsuperscript{1}Warsaw University of Technology,
 \textsuperscript{2}Wroclaw Tech,
 \textsuperscript{3}University of Wroclaw
\\
   \href{wiktoria.kowszewicz@gmail.com}{wiktoria.kowszewicz@gmail.com, kazienko@pwr.edu.pl}
}

\begin{document}
\maketitle
\begin{abstract}
In this work we present the Social Influence Technique Taxonomy (SITT), a comprehensive framework of 58 empirically grounded techniques organized into nine categories, designed to detect subtle forms of social influence in textual content. We also investigate the LLMs ability to identify various forms of social influence. Building on interdisciplinary foundations, we construct the SITT dataset -- a 746-dialogue corpus annotated by 11 experts in Polish and translated into English -- to evaluate the ability of LLMs to identify these techniques. Using a hierarchical multi-label classification setup, we benchmark five LLMs, including GPT-4o, Claude 3.5, Llama-3.1, Mixtral, and PLLuM. Our results show that while some models, notably Claude 3.5, achieved moderate success (F1 score  = 0.45 for categories), overall performance of models remains limited, particularly for context-sensitive techniques. The findings demonstrate key limitations in current LLMs' sensitivity to nuanced linguistic cues and underscore the importance of domain-specific fine-tuning. This work contributes a novel resource and evaluation example for understanding how LLMs detect, classify, and potentially replicate strategies of social influence in natural dialogues.
\end{abstract}
\section{Introduction}
In recent years, large language models (LLMs) have demonstrated remarkable proficiency in understanding and generating human-like text \cite{chang2024survey}. 
As these systems become increasingly embedded in domains with significant public influence -- such as journalism \cite{simon2024artificial} and politics \cite{tretter2025opportunities} -- their ability to grasp not only what is said but how it is said -- or what is left unsaid -- becomes critically important.
LLMs have shown strong capabilities to detect disinformation, highlighting their proficiency in pattern recognition and contextual understanding \cite{sosnowski2024eu,kuntur2024under,papageorgiou2024survey}. However, beyond surface-level misinformation, the more subtle dimensions of language -- such as rhetorical framing, omissions, or social influence techniques -- pose new challenges, including difficulties in detection, interpretation of intent, and distinguishing persuasion from manipulation. These nuances can significantly shape people's perception, beliefs,  emotions, attitudes, and finally behavior.

To better leverage the potential of LLMs in detecting various forms of harmful content, we developed a two-tier taxonomy of social influence categories and techniques and evaluated LLMs' ability to identify them. This study is grounded in the theoretical distinction between social influence -- the broad set of processes by which individuals’ attitudes, beliefs, or behaviors are shaped by others \cite{cialdini2021}; persuasiveness -- the strategic, often transparent use of communication and influence to change people's attitudes or actions \cite{perloff1993dynamics}; and manipulation -- a covert or deceptive form of influence that bypasses rational scrutiny or autonomy  \cite{buss1992manipulation}. 

The main aim of this study is to evaluate the ability of LLMs to identify instances of social influence in real-life short conversational texts, using a newly developed taxonomy of social influence. 
In this study, we address the following research questions:\\
\textbf{[R1]} How to group the techniques of social influence used in textual communication into broader categories?\\
\textbf{[R2]} How effective are LLMs in identifying social influence categories and techniques? 

This work makes the following contributions: \\
\textbf{[C1]} We introduce a novel taxonomy of 58 techniques grouped into nine categories of social influence, along with their definitions and examples. \\
\textbf{[C2]} We present a new annotated dataset with social influence techniques.\\
\textbf{[C3]} We test the performance of LLMs in recognizing social influence  techniques.



\section{Related Work}
\subsection{Social influence in the social sciences: key mechanisms}
Social influence is a fundamental aspect of human behavior and plays a critical role in shaping interpersonal interactions. People often adjust their beliefs, preferences, or actions in response to social influence, whether consciously or unconsciously. In the domain of social communication, most recognized methods by which an individual influences other persons’ decisions and behavior include various principles, such as commitment and consistency, social proof, reciprocity, liking and sympathy, scarcity, authority developed by \citet{cialdini2021}. Kahneman’s work informs how heuristics, framing, and loss aversion shape social influence \cite{kahneman2011fast}. In their comprehensive review, \citet{dolinski2022100} further organized experimental studies into categories such as emotional appeals, sequential strategies, interpretive frameworks, and identity-based mechanisms.

Building on these conceptual foundations, we developed our own taxonomy of social influence and persuasion techniques specifically tailored for studying how LLMs detect and interpret instances of social influence in textual communication. Our taxonomy integrates various dimensions of persuasion to optimize detection performance in AI-driven language analysis.

\subsection{Social influence taxonomies in computer science}

Although the main theoretical foundations stem from social and behavioral research, computer science also provides custom taxonomies and definitions frequently developed along with its own datasets. 

Two studies \cite{El-Sayed2024, Jones2024} distinguished between manipulation and persuasion, although in some cases persuasion is used as an umbrella term that includes both manipulation and `rational persuasion`. Categorization within the area of persuasion has received much more attention, with several small-scale taxonomies, typically covering no more than 10 techniques \cite{Ma2025, wang2020persuasiongoodpersonalizedpersuasive}, with the exception of one larger taxonomy \cite{zeng2024johnnypersuadellmsjailbreak}, however lacking textual resources. Additionally, \cite{Kumar2023} introduced a taxonomy of persuasive techniques in advertisements, which is focused on image-based features.

Although studies are limited, efforts have already been made to develop a unified taxonomy \cite{Oyibo2024}, driven by inconsistencies in existing approaches. 
\subsection{LLMs in recognizing social influence}

Social influence research has primarily focused on detecting persuasion. Several datasets \cite{piskorski-etal-2023-semeval, jin-etal-2024-persuading} and data generation methods \cite{Tiwari2023, Zhang2025} have been proposed, though they typically rely on small-scale taxonomies. A limited number of studies addressed detection of manipulative or persuasive techniques \cite{wang2024mentalmanipdatasetfinegrainedanalysis}, but comprehensive research on social influence as a broader phenomenon remains scarce. The majority of works used language models to conduct only preliminary experiments, without proposing any advanced methodologies.
As an exception, \cite{Singh2024} introduced a concept of `transuasion` -- transformation of non-persuasive content into persuasive one, aimed at generating persuasive material, particularly for advertisements. Further research may contribute to the development of automated systems for detecting disinformation and deceptive persuasive tactics, supporting ethical communication management across domains such as organizations.

\section{Social Influence Techniques Taxonomy (SITT)}
\label{sec:SITT}
Based on the most representative summaries, classifications, and main techniques of social influence \cite{dolinski2022100, cialdini2021, kahneman2011fast}, we propose an original taxonomy of techniques that includes the most well-known, empirically verified mechanisms of social influence. The Social Influence Technique Taxonomy (SITT) primarily draws upon the systematization of techniques presented in a review by \cite{dolinski2022100}, supplemented by four techniques referring to Cialdini's  rules \cite{cialdini2021}, and the framing effect \cite{kahneman2011fast}. The main criteria for selecting techniques for the SITT were their relevance and applicability to textual content, as well as their detectability by LLMs.

Nine expert judges conducted a qualitative semantic analysis of various techniques to identify common mechanisms of human influence based on their definitions. This process produced a distinct set of 58 techniques, grouped into nine content categories, Appendix~\ref{Social Influence Technique Taxonomy (SITT) - definitions and examples of techniques}. Each category reflects shared social influence mechanisms, while the techniques themselves remain relatively independent. Table~\ref{tab:persuasion-techniques} shows the SITT taxonomy with definitions of social influence categories and the respective  techniques.

\begin{table*}[h!]
\centering
\small
\begin{tabular}{
    >{\raggedright\arraybackslash}p{3.3cm}
    >{\raggedright\arraybackslash}p{5.2cm}
    >{\raggedright\arraybackslash}p{6.2cm}
}
\toprule
\textbf{Social influence category} & \textbf{Definition} & \textbf{Social influence techniques} \\
\midrule

A. Appeal to a positive or negative image (\textit{Image})& 
Refers to the recipient's self-evaluation in terms of identity, dignity, morality, or social image – to induce behavior consistent with a positive image or avoiding a negative one. & 
1. Expert snare 2. To be exceptional 3. You will probably refuse, but\dots 4. Labeling  5. A witness to an interaction 6. We are looking for people like you\\
\midrule

B. Context modification (\textit{Context})& 
Refers to modifying or using elements of context (situation, time, place, or space) to influence perception. The information remains the same, but its presentation context changes. & 
7. Framing 8. Disrupt-then-reframe 9. Ask for it well in advance 10. Face-to-face meeting 11. Unavailability 12. Goal progress 13. The power of limited choice\\
\midrule
C. Biased presentation of information and/or arguments (\textit{Information})& 
Refers to presenting information in biased, selective, suggestive, ambiguous or simplified ways, distorting the message to evoke specific opinions or decisions. & 
14. Dump and chase 15. Script of mindless action 16. Validation–persuasion 17. Induction of hypocrisy 18. Valence framing 19. Pique technique 20. The only request\\
\midrule

D. Appeal to social consensus and group norms \textit{(Social norms)}& 
Refers to inducing behaviors by referencing majority behavior or social norms. & 
21. Metacommunication bind 22. Everyone knows it 23. The “We” rule 24. That’s how we do it here 25. We are exceptional\\
\midrule

E. Appeal to social reciprocity (\textit{Reciprority})& 
Refers to basic human feeling of obligation to reciprocate benefits or favors. & 
26. Birthday effect 27. Gratitude 28. Give to take 29. Indirect reciprocity 30. We've already given 31. Door-in-the-face\\
\midrule

F. Appeal to emotions (\textit{Emotions})& 
Refers to deliberately evoking emotional states (fear, guilt, joy, disappointment) to trigger behaviors or thoughts. & 
32. Emotional see-saw 33. Fear and anxiety 34. Anticipatory regret 35. Take advantage of good mood 36. Take advantage of bad mood 37. Physiological arousal 38. Guilt 39. Shame 40. Embarrassment 41. Show disappointment 42. Positive cognitive state 43. Humor 44. Foot-in-the-mouth 45. The power of word “love” 46. Cognitive exhaustion\\
\midrule

G. Appeal to sympathy, liking, connections \textit{(Liking)}& 
Refers to creating sympathy or emotional closeness to increase persuasiveness. & 
47. Liking 48. Similarity 49. Flattery\\
\midrule

H. Appeal to authority \textit{(Authority)}& 
Refers to knowledge, social position, or credentials (e.g., science, titles, institutions) to boost argument credibility. & 
50. Authority of person or science\\
\midrule

I. Appeal to consistency in views and/or behavior \textit{(Consistency)}& 
Refers to invoking the human need to maintain consistency in beliefs and behaviors. & 
51. That's not all 52. Default settings 53. Inducing commitment 54. Low ball 55. Foot-in-the-door 56. Four walls 57. Even a penny or moment will help 58. Make your commitments public\\
\bottomrule
\end{tabular}
\caption{SITT categories and techniques}
\label{tab:persuasion-techniques}
\end{table*}

\section{The SITT dataset}

We present the SITT Dataset\footnote{\url{https://github.com/social-influence/sitt-dataset/}}, containing 746 human-annotated dialogues based on the developed taxonomy. The annotations were performed in Polish, and additional English translations of the dialogues were created using GPT-4o, Appendix~\ref{Prompt:translate_from_polish}. The final set of categories and techniques combines inputs from all annotators.

\subsection{Dialogue corpus}
\label{sec:corpus}
The corpus was constructed using dialogues from three sources: (1) a random sample of 488 instances from the MentalManip dataset \cite{wang2024mentalmanipdatasetfinegrainedanalysis}, each containing an identified social influence technique; (2) 99 persuasive samples from the evaluation set of the CToMPersu dataset \cite{Zhang2025}; (3) 159 samples generated by GPT-4o -- 3 samples per technique based on the developed taxonomy, Appendix~\ref{Prompt:generation}. As some examples were poorly created, they were manually removed from the corpus, resulting in a final sample of 746 dialogues. The above values have been manually refined. 

The data were processed through the following steps. First, the samples from MentalManip and CToMPersu were translated into Polish, Appendix~\ref{Prompt:translate}, and then post-edited to highlight potential instances of social influence in the text, Appendix~\ref{Prompt:enhance}. Subsequently, the processed texts were pre-assigned to SITT categories using GPT-4o, Appendix~\ref{Prompt:Category_Label}, and finally verified by experts.

\subsection{Expert-based dialogue annotation}

The resulting corpus was annotated by 11 experts who completed a total of 2,177 assignments. The experts were recruited among graduate students and working professionals with expertise in social influence techniques. Nine of them were women and two were men, with an average age of 23.18.  To minimize cognitive overload and ensure annotation quality, each expert was assigned two categories of social influence. Each annotator was given a fair salary for their work in the amount of 1500 PLN. This research received suitable approval no. O-25-10 from the Ethics Committee. 

Annotation samples were distributed based on initial model predictions, ensuring that each predicted category for a given sample was assigned to two annotators. 
For instance, a sample with predicted categories \([2, 3, 6, 8]\) might be reviewed by four annotators -- two responsible for categories \(2, 3\), and two for \(6, 8\). The total number of annotators per sample could be higher, depending on the assigend category. The actual combinations that were used included two annotators for each of the following category sets: \(\{6, 8\}, \{2, 5\}, \{1, 7\}, \{9, 4\}\), and three annotators for the set \(\{3, 6\}\). The choice was based on the label distribution.

Annotators completed two tasks: (1) identifying and labeling sentences that contained social influence techniques within their assigned categories, and (2) marking the presence of other categories of social influence. To mitigate task-order bias, half of the annotators performed category selection first, followed by technique identification, while the other half began with technique selection. The annotation process was conducted using Argilla \cite{vila-suero2023argilla}.

There were also 5,378 sentence-based annotations not yet verified by experts, which will be used in the future explanatory task. 
\subsection{Expert verification}
The goal of this step was to ensure the quality of annotations. Three domain experts verified a minimum of 10\% of each annotator's annotations, checking whether their decisions were consistent with the technique definition. Correctness of annotators varied from 73\% to 100\% (Mean = 87\%), Appendix~\ref{sec:expert-verification}. 

\subsection{Dataset profile}
\begin{figure}[ht]
    \centering
    \includegraphics[trim={0.4cm 0.5cm 0.4cm 1cm}, clip, width=1\columnwidth]{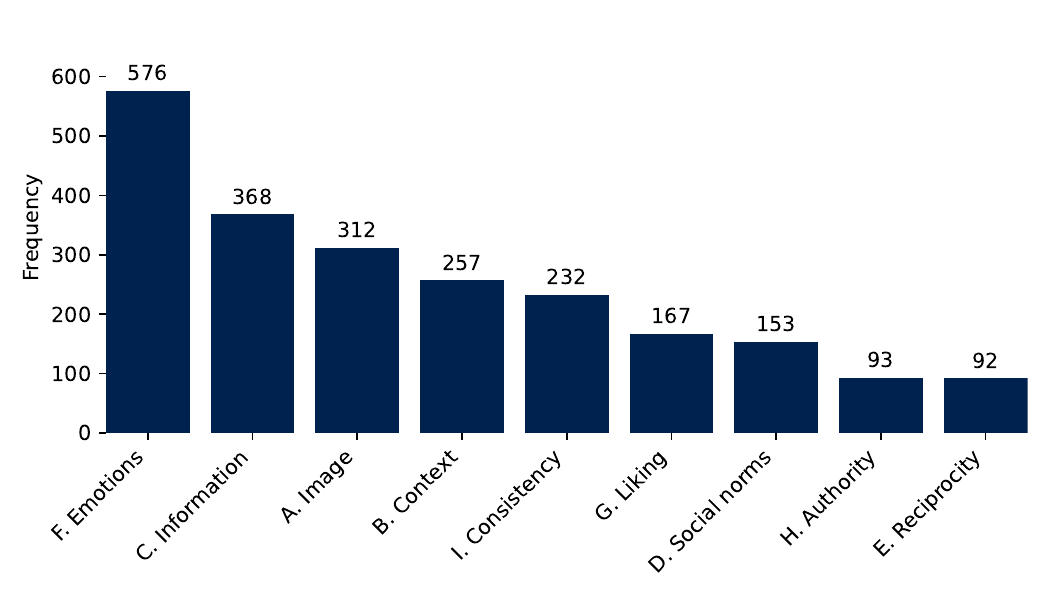}
    \caption{The frequency of annotated categories in the whole SITT dataset.}
    \label{fig:categories} 
\end{figure}

\begin{figure*}[ht]
    \centering
    \includegraphics[trim={0 0.5cm 0 0.2cm},clip,width=\textwidth]{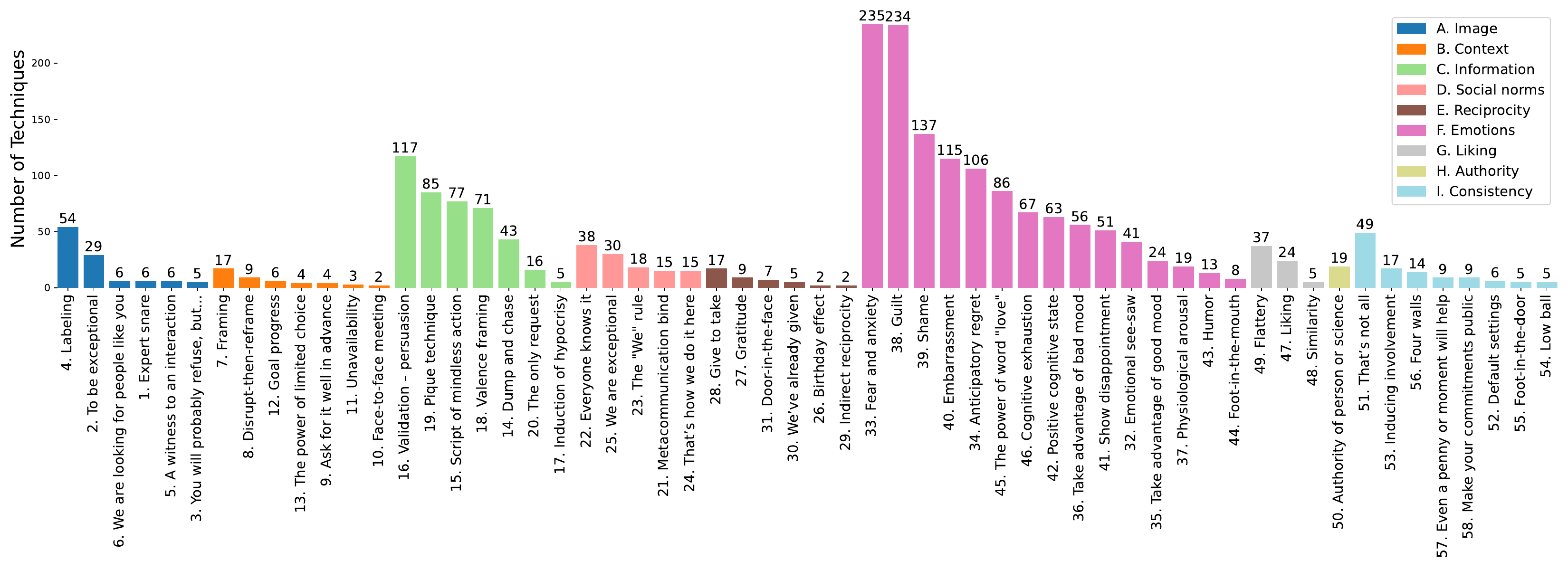}
    \caption{Unique expert-verified occurrences of SITT techniques in all dialogues.}
    \label{fig:techniques} 
\end{figure*}

The resulting SITT dataset contains 746 dialogues with an average of 6.46 turns (SD = 5.45) annotated with nine categories and 58 techniques (Section~\ref{sec:SITT}), Figure~\ref{fig:categories} and~\ref{fig:techniques} for details about class distributions. The most numerous category, appearing in almost 77\% of dialogues, was the \textit{Appeal to emotions}, with techniques like \textit{Fear and anxiety, Guilt, or Shame} being the most frequently used.

\begin{figure}[ht]
    \centering
    \includegraphics[trim={0.6cm 0 1.7cm 0.8cm}, clip, width=1\columnwidth]{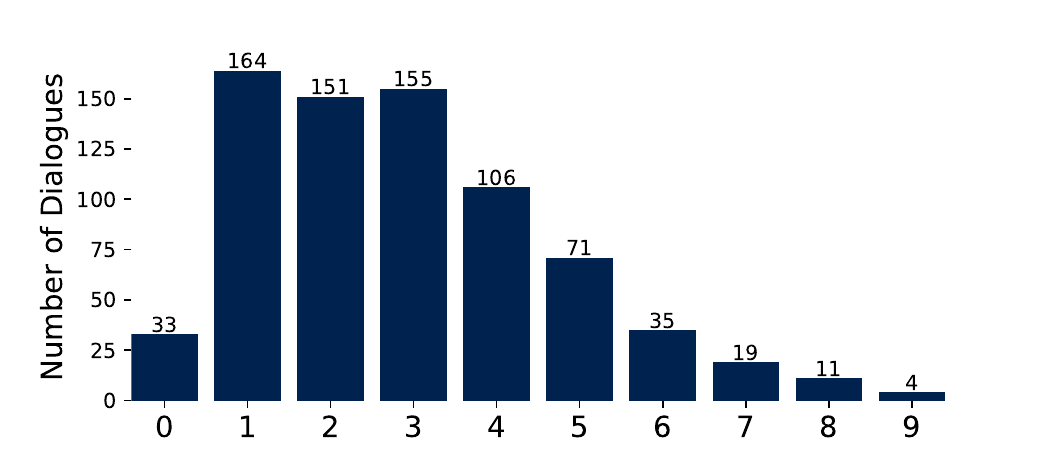}
    \caption{The distribution of unique SITT techniques per dialogue.}
    \label{fig:distribution} 
\end{figure}

On average, each dialogue was annotated with 2.9 techniques (SD = 1.8, Figure~\ref{fig:distribution}) and 3 categories (SD = 1.5). There were only 43 dialogues with no assigned technique. As annotators were tasked to mark all categories they recognize, but only those techniques they specialize in, the mean number of categories per dialogue is higher, despite the possibility of many techniques from a single category. This is because the second round of annotations, in which such samples would be assigned to appropriate technique annotators, was not yet performed.

\begin{figure}[ht]
    \centering
    \includegraphics[trim={0.2cm 0.3cm 0 0cm}, clip, width=1\linewidth]{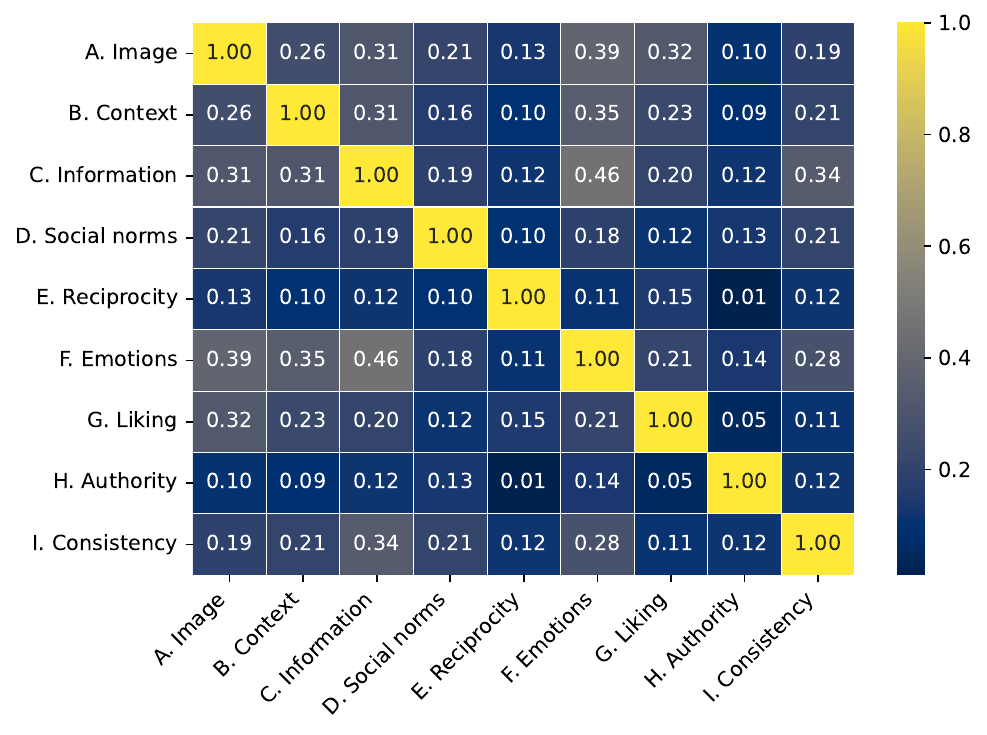}
    \caption{The co-existence of categories in dialogues using the Jaccard score.}
    \label{fig:jaccard}
\end{figure}

We calculated the coexistence of categories in dialogues with the Jaccard metric \cite{jaccard1901etude}, Figure~\ref{fig:jaccard}. The most frequent co-occurrence between \textit{Appeal to emotions} and \textit{Biased presentation of information and/or arguments} was 0.46, while the average overall co-occurrence was 0.17. 

\section{Social influence detection with LLMs}
\subsection{Experimental setup}
Below we present the details of the experimental setup of our research.

\textbf{Models}. 
We selected five different LLMs (Table~\ref{tab:models}) to evaluate their detection capabilities of the categories and techniques included in SITT. Some LLMs were commercial, closed-access (Claude, GPT), whereas some others -- open-weights like Llama and Mixtral. For Polish dialogues only, we also tested PLLuM, the largest Polish-specific model.

\textbf{Hierarchical classification}.
The classification of the SITT categories and techniques was carried out in a hierarchical manner. In the first step, the models were solely queried about which SITT categories they would assign to a given dialogue. If the model response was difficult for us to interpret, the absence of social influence techniques was assumed. Subsequently, the model was asked to identify specific techniques from the SITT list, but only those associated with the categories previously predicted by the model. 

\textbf{Prompts}. 
To obtain data in English, it was necessary to translate the dataset. For this purpose, the prompt provided in Appendix \ref{Prompt:translate_from_polish} was used. From this point onward, the only differences in the classification processes concerned the language of the data and the prompts employed.

The prompt used for categories from the SITT began with an instruction on how to assess the text, Appendix~\ref{Prompt:categories_classification}. This was followed by a presentation of all the categories, each accompanied by definitions and examples. The expected response format was then outlined. Additionally, the model was instructed that if no instance of social influence was present in the text, it should indicate what would need to be changed in the text for such influence to appear. Finally, the text to be evaluated was presented. 

The second prompt was designed to guide the models in the task of classifying the technique from the SITT list, Appendix~\ref{Prompt:technique_classification}. It began with a task description, followed by a presentation of only those techniques, with definitions and examples, that corresponded to the previously classified categories. Then, it includes the expected response format. In addition, the model was asked to provide an explanation for its selection of techniques. The evaluated text was presented at the end. 

\textbf{Parameters}. 
For the classification task, we used model parameters with \textit{temperature} set to 0.0 and the \textit{top\_p} parameter also set to 0.0.

\begin{table}[ht]
\scriptsize
\centering
\begin{tabular}{llll}
\toprule
\textbf{Model} & \textbf{Language} & \textbf{References} \\ \midrule
gpt-4o-2024-08-06 & PL, EN & \citet{openai2024gpt4ocard} \\ 
claude-3-5-sonnet-20240620 & PL, EN & \citet{claude35sonnet2024} \\ 
Mixtral-8x22B-Instruct-v0.1 & PL, EN & \citet{mixtral8x22b2024} \\ 
Meta-Llama-3.1-70B-Instruct & PL, EN & \citet{meta2024llama3.1} \\ 
PLLuM-8x7B-nc-chat & PL & \citet{pllum2025} \\
\bottomrule
\end{tabular}
\caption{Large Language Models used in experiments.}
\label{tab:models}
\end{table}





\subsection{Results}
The scores of LLMs in multi-label classification of both SITT categories and techniques are shown in Table~\ref{tab:models-f1}. Each reported value corresponds to the micro-averaged F1 score. For Polish, Claude performed best in all metrics, both for category and technique assessment, achieving 0.45 and 0.31 F1 score for categories and techniques respectively. For English, Claude remains best in techniques, achieving 0.29 F1 score, but Mixtral comes on top in categories with 0.4 F1 score. More detailed per-category and per-technique results are presented in Appendix~\ref{additional-results}.

\begin{figure*}[]
    \centering
    \includegraphics[width=1\linewidth]{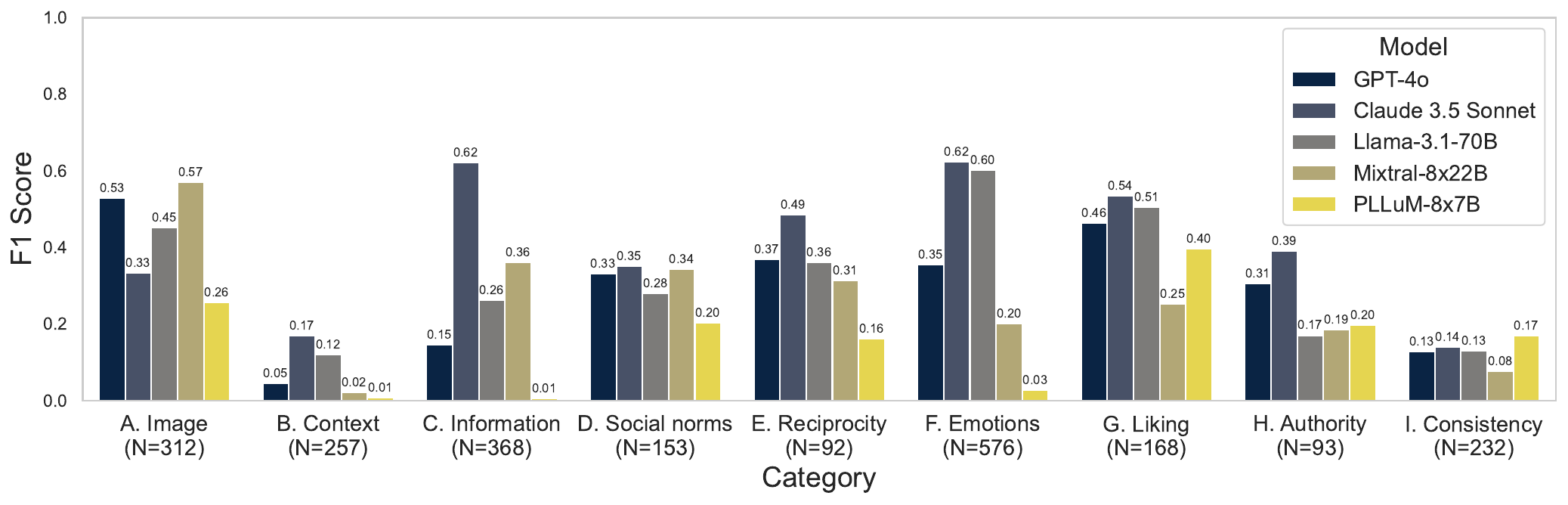}
    \caption{F1 scores of tested LLMs for classifying SITT categories.}
    \label{fig:f1_score_categories}
\end{figure*}

The performance of each model for the SITT categories is presented in Figure \ref{fig:f1_score_categories}. It reveals substantial differences between categories and model performance. With respect to the most populous classes in the dataset, Claude 3.5 achieved the highest F1 score of 0.62 in both \textit{Emotions} and \textit{Information} categories. In the \textit{Image} category, the best performance was attained by Mixtral-8x22B, with an F1 score of 0.57. The lowest overall result was observed in the \textit{Context} category, where the mean F1 score reached only 0.07.

\begin{figure*}[h!]
    \centering
    \includegraphics[width=1.0\linewidth]{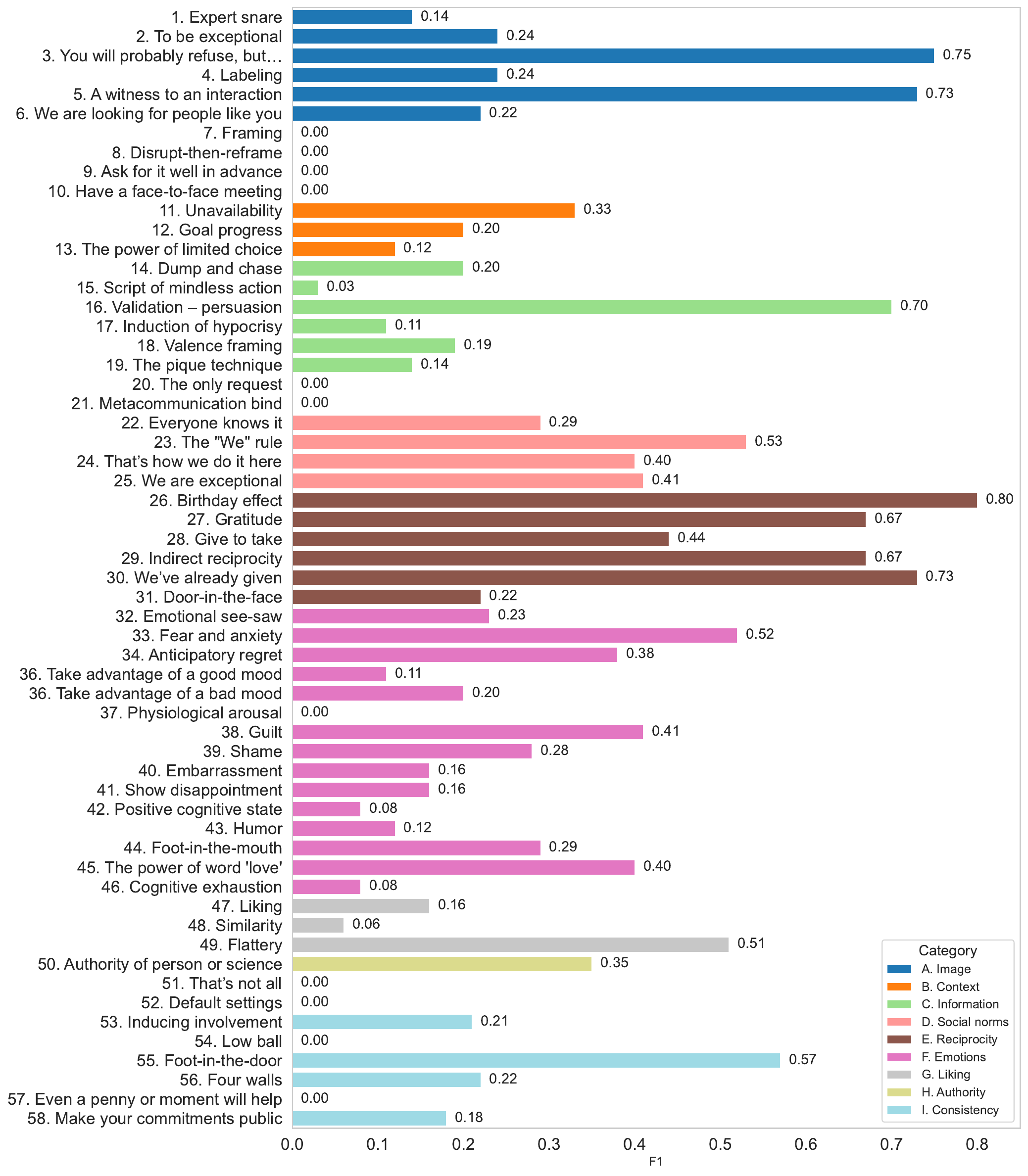}
    \caption{F1 scores of Claude Sonnet 3.5 for classification of the SITT techniques based on Polish dataset version.}
    \label{fig:techniques_claude_PL}
\end{figure*}

Figures \ref{fig:techniques_claude_PL} and \ref{fig:appendix_techniques_claude_EN} illustrate the performance of Claude 3.5 Sonnet across all SITT techniques. The results reveal that 11 techniques in the Polish dataset and 14 in the English dataset were never correctly identified, as they received no correct annotations. For the Polish data, the highest F1 score (0.8) was achieved for technique 26. \textit{the Birthday Effect}.


\begin{figure}[]
    \centering
    \includegraphics[width=1\linewidth]{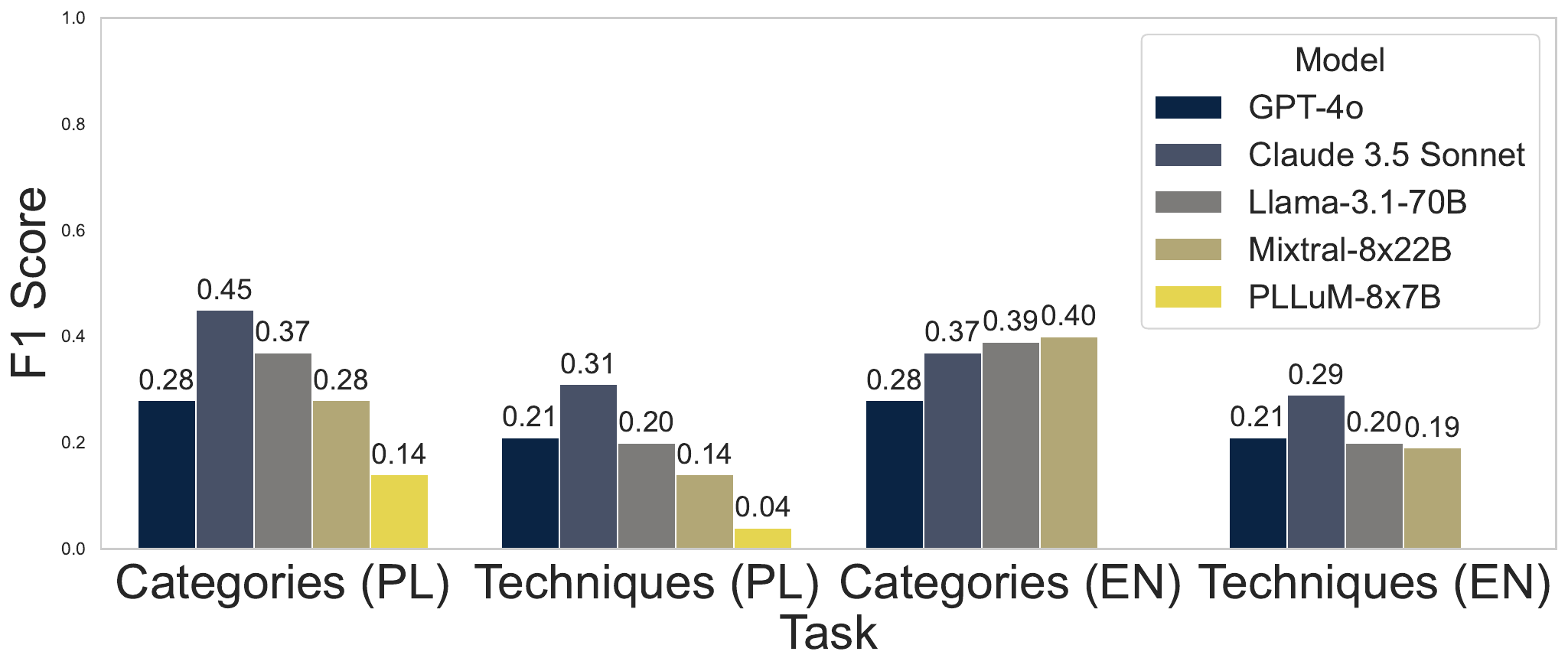}
    \caption{F1 scores of tested LLMs for classifying SITT categories and techniques.}
    \label{fig:f1_score}
\end{figure}

\begin{table*}[ht]
\centering
\small
\begin{tabular}{lSSSSSSSS}
\toprule
\multirow{2}{*}{\textbf{Model}} & \multicolumn{4}{c}{\textbf{Categories}} & \multicolumn{4}{c}{\textbf{Techniques}} \\
\cmidrule(lr){2-5} \cmidrule(lr){6-9}
& \textbf{F1} & \textbf{Precision} & \textbf{Recall} & \textbf{Jaccard} & \textbf{F1} & \textbf{Precision} & \textbf{Recall} & \textbf{Jaccard} \\
\midrule
GPT-4o (PL) & 0.280 & 0.505 & 0.194 & 0.163 & 0.213 & \bfseries 0.43 & 0.142 & 0.119 \\
Claude 3.5 Sonnet (PL)  & \bfseries 0.446 & \bfseries 0.647 & \bfseries 0.340 & \bfseries 0.287 & \bfseries 0.311 & \bfseries 0.430 & \bfseries 0.247 & \bfseries 0.19 \\
Llama-3.1-70B (PL) & 0.371 & 0.604 & 0.268 & 0.228 & 0.202 & 0.299 & 0.153 & 0.113\\
Mixtral-8x22B (PL) & 0.278 & 0.460 & 0.199 & 0.161 & 0.144 & 0.211 & 0.109 & 0.077\\
PLLuM-8x7B (PL) & 0.144 & 0.232 & 0.104 & 0.077 & 0.037 & 0.076 & 0.02 & 0.019 \\
\midrule
GPT-4o (EN) & 0.279 & 0.495 & 0.194 & 0.162 & 0.207 & \bfseries 0.419 & 0.137 & 0.115 \\
Claude 3.5 Sonnet (EN)  & 0.374 & 0.592 & 0.274 & 0.230 & \bfseries 0.287 & 0.369 & \bfseries 0.234 & \bfseries 0.167 \\
Llama-3.1-70B (EN) & 0.388 & \bfseries 0.618 & 0.283 & 0.241 & 0.200 & 0.292 & 0.153 & 0.111 \\
Mixtral-8x22B (EN) & \bfseries 0.396 & 0.570 & \bfseries 0.304 & \bfseries 0.247 & 0.194 & 0.218 & 0.17 & 0.107\\
\bottomrule
\end{tabular}
\caption{LLM performance in detecting social influence categories and techniques in Polish and English.The best result in each column is marked in bold. All metric aggregations are micro aggregations.}
\label{tab:models-f1}
\end{table*}

\section{Discusion} 
The purpose of this paper was twofold: first, to explore how social influence techniques can be meaningfully grouped into broader categories; and second, to evaluate the ability of LLMs to detect these categories and techniques in real-life conversational texts. Our findings show that, while generally LLMs can detect some categories and techniques quite well, they perform poorly in identifying many others. None of the LLMs achieved an F1 score above 0.62 for any category, Figure~\ref{fig:f1_score_categories}. Claude 3.5 Sonnet appeared to perform best. In general, the results reveal that the available LLMs require additional training (fine-tuning) to achieve higher expertise in detecting many more techniques.
It was also observed that, in some cases, the models were able to identify a category but failed to assign a corresponding technique.

The investigated models  were generally cautious when classify text as persuasive or manipulative, but if they do such classifications, they were highly accurate. This accuracy minimizes false alarms and reduces the risk of unjustified accusations related to the use of social influence techniques. However, the low recall rate suggests that the models overlooked a substantial amount of manipulative content. This may protect people from excessive control or interference with freedom of speech if LLMs are utilized as monitors.

GPT 4o results were identical for Polish and English samples, probably because this model was used for the translation task.

The annotators also marked text units (sentences) as evidence of a given technique. However, it seems obvious that annotated texts still require a resource-intensive and expensive verification. 
Once corrected, they will be used to fine-tune the LLM not only to recognize social influence techniques but also to precisely explain it to the user. 

\section{Conclusions and future work}
In conclusion, we introduced a taxonomy of 58 social influence techniques, systematically organized into nine categories (SITT). This taxonomy is grounded in both theoretical frameworks and empirical findings from the social sciences. Using it, we experimentally demonstrated that LLMs have not yet developed strong capabilities for detecting social influence categories and techniques. Their performance varied substantially across different categories, suggesting that certain forms of social influence remain particularly challenging for LLMs to detect. Improving their performance most probably requires larger datasets.

Claude model showed the strongest capabilities in detecting the social influence categories and techniques in the Polish language. This suggests that LLMs may have language-specific strengths in psychological domains. It emphasizes the importance of evaluating model behavior not just across tasks but also across languages individually.
We noticed that the results in Polish were, on average, higher than those in English, which may be explained by the fact that the data were originally annotated in Polish and then translated into English.
Interestingly, the performance of GPT-4o was average compared to the other models. However, it is worth noting that this model achieved identical results for both Polish and English texts. This outcome is likely due to the fact that the second round of annotations—during which samples would be assigned to annotators for specific techniques—had not yet been conducted.

The models had a problem with classifying \textit{Context} and \textit{Consistency} SITT categories. There are many SITT techniques within these categories that models were unable to classify even once, Figure~\ref{fig:techniques_claude_PL}.
This may suggest that the models have difficulty detecting categories that require contextual information—whether situational or personal. 
The context modification category requires more information that humans intuitively deduce based on their internal knowledge. On the other hand, consistence category demands an understanding of the personal context which humans intuitively understand. 

Our future work will focus on (1) extension of the SITT corpus to increase the quantity of underrepresented techniques, (2) fine-tuning of LLMs to boost their reasoning capabilities, (3) exploiting sentence-based annotations to train models that explain and show a particular manipulations to the user, and (4) respect diverse contexts during inference related to social influence detection.

\section*{Limitations}
Several limitations of the present study should be acknowledged.

First of all, as this was a pilot study, no calibration session was conducted to address potential uncertainties encountered during annotations. Therefore, in future work, we plan to run regular calibration sessions to improve consistency. Additionally, only a single annotation attempt was carried out.

Secondly, many of the proposed techniques can be used in other types of texts, not only dialogues. We also acknowledge that social influence detection from text data may not capture its complexity, with factual user emotions, current mood, and other modalities of communication.

Next, the effectiveness of a technique classification is strongly dependent on the degree to which the models recognize the overarching category. Thus, misannotations at the category level prevented the models from identifying the correct detailed techniques.

Also, challenges in data acquisition and generation constrained the diversity of texts used. The analysis was limited to a specific set of language models; future work should include additional commercial systems.

As a preliminary step, we employed a language model instead of human annotators to assign texts to categories of social influence. Although in our study the annotators assigned categories to texts, this occurred simultaneously with the annotation of specific techniques. The resulting dataset is expected to be more suitable for the second phase of the study.

Another limitation is that the annotators were not demographically diversed in terms of age and sex, potentially limiting the range of perspectives in the annotation process. 
Furthermore, the SITT dialogue dataset was highly imbalanced in techniques, Figure~\ref{fig:techniques}, due to the lack of many techniques in the MentalManip and CToMPersu component sets, Section~\ref{sec:corpus}. Unfortunately, we were unable to recognize this before completing the annotations. The next edition of the corpus will address this issue.

AI assistants were used solely for linguistic support, and we recognize the potential risks of misuse when applying LLMs for influence detection in sensitive contexts.

\bibliography{main}
\clearpage

\appendix
\onecolumn

\noindent\begin{minipage}{\textwidth}
  \begin{multicols}{2}
    \section{Social Influence Technique Taxonomy (SITT) - definitions and examples of techniques}
\label{Social Influence Technique Taxonomy (SITT) - definitions and examples of techniques}
    \addcontentsline{toc}{section}{Long Table Section}
  \end{multicols}
\end{minipage}

\captionsetup[longtable]{justification=centering, width=\textwidth}
\begin{longtable}{p{5.5cm}p{8.5cm}}
\toprule
\textbf{Name/Definition} & \textbf{Example} \\
\midrule
\endfirsthead

\midrule
\textbf{Name/Definition} & \textbf{Example} \\
\midrule
\endhead

\multicolumn{2}{r}{\small Continued on the next page} \\
\endfoot

\endlastfoot

\multicolumn{2}{l}{\textbf{A. Appeal to a positive or negative image \textit{(Image)}}} \\
\midrule
\textit{1. Expert snare} \newline
Emphasizing the interlocutor's expertise during the conversation encourages him to maintain this image and act in accordance with the assigned role. & "It is clear that you have a deep understanding of animals and I believe that you will find that our product is an excellent solution for cats with sensitive stomachs." \\
\midrule
\textit{2. To be exceptional} \newline
Emphasizing the uniqueness of a person/group makes them more inclined to engage in a particular behavior. & "But you are extraordinary. And that is why you (and only you) will be treated in a special way."\\
\midrule
\textit{3. You will probably refuse, but\ldots} \newline
Suggesting to the interlocutor that she is likely to refuse to comply with the request, which paradoxically leads her to accept it. & "You will probably refuse, but I wonder if you would be willing to help us by making a monetary donation." \\
\midrule
\textit{4. Labeling} \newline
Presenting the interlocutor's characteristics in a way that makes her believe that the characteristics are true. As a result, the person behaves consistently with the described characteristics. & "You are the head of the family. You will definitely make the decision that is best for our family." \\
\midrule
\textit{5. A witness to an interaction} \newline
Using the presence of a witness to induce the interlocutor to make a decision -- to grant or reject the request -- in a way that reinforces her desired image. & Kasia and Tomek are walking together in the market square. They are approached by a volunteer collecting donations for an animal shelter. Tomek, wanting to make a good impression on Kasia, takes out his wallet and pays 50 zlotys. \\
\midrule
\textit{6. We are looking for people like you}\newline
Asking a person while emphasizing that they are looking for someone with specific characteristics that the person has. & "I'm looking for people who really care about the environment, like you,  which increases the chances of receiving a donation. "Would you like to support our action with a small donation?" \\
\midrule
\multicolumn{2}{l}{\textbf{B. Context modification} (\textit{\textbf{Context)}}} \\
\midrule
\textit{7. Framing} \newline
Presenting information in a specific context or "framework" that influences the way people interpret it and make decisions. & "This treatment has a 90\% success rate" (positive framing). "There is a 10\% risk of surgery failure" (negative framing). \\
\midrule
\textit{8. Disrupt-then-reframe} \newline
Putting a person in a state of confusion or uncertainty that makes them less able to rationally analyze the situation. & A salesperson presents a customer with several different phone models, each time changing their opinion about them (e.g., "This model is the latest, but this one has a better camera, and this one is more functional"). Confused and indecisive, the customer becomes more susceptible to the persuasion of the seller. \\
\midrule
\textit{9. Ask for it well in advance} \newline
Asking to take action well in advance, as people rate their future responsibilities as less burdensome than their current ones. & An industry conference organizer invites you to speak in eight months. Because the event seems far off, you agree without hesitation, assuming that you will have plenty of time to prepare. However, as the deadline approaches, your schedule is packed — and backing out is no longer an option. \\
\midrule
\textit{10. Have a face-to-face meeting} \newline
Encouraging direct contact, making it easier to build trust and increase the chance of a positive response. & A: "Thank you for taking the time. I have a request for you -- could you help me with the preparation of the report? Your experience would be very valuable to me." B: "I understand, I'll be happy to help." \\
\midrule
\textit{11. Unavailability} \newline
Assigning more value to things that are more difficult to access or are limited in time and quantity. Generating a sense that something is special and valuable increasing the desire to possess it. & Promotion lasts only until July 14. Hurry! The number of products included in the promotion is limited. \\
\midrule
\textit{12. Goal progress} \newline
In the message, emphasizes that the achievement of the goal is imminent, so that the person continues to action. & "Look ahead and see how close you are. You will travel a few hundred meters more and you are at the top." \\
\midrule
\textit{13. The power of limited choice} \newline
The purpose of this is to steer an individual in the desired direction by limiting the number of options available to choose from. & "Get involved in saving our planet! As part of our campaign, you can: a) plant a tree in a designated place or b) pay PLN 20 to buy a seedling. By choosing one of the two ways, you can help returning to nature what we have taken from it. Only together can we act effectively!" \\
\midrule
\multicolumn{2}{l}{\textbf{C. Biased presentation of information and/or arguments \textit{(Information)}}} \\
\midrule
\textit{14. Dump and chase}\newline
After an obstacle to implementing a request arises, the continuation of the dialogue by asking questions to explain the reasons for the refusal. & The caller refuses due to a lack of time. We can propose a different date or a shorter meeting, which increases the chance of acceptance. \\
\midrule
\textit{15. Script of mindless action} \newline 
Adding any (even trivial) justification to a request. & A: Will we postpone the deadline for submitting the project by one day? B: That can be problematic. A: An extra day will allow you to gather all the necessary information. \\
\midrule
\textit{16. Validation - Persuasion}\newline
Admitting that the interlocutor is right that his resistance to change or action is understandable, and then presenting arguments to convince him to take the desired action. 
&
A dietitian tells a person who is losing weight: "I know how difficult it is to give up chocolate because it is delicious and improves your mood, but to improve your health and avoid diabetes, it is worth making changes to your diet." \\
\midrule

\textit{17. Induction of hypocrisy}\newline
To obtain statements from a person in support of certain attitudes or behaviors, and then to demonstrate that their actions are contrary to those statements. 
&
A: I know smoking is harmful; everyone knows that. B: Then why do you still smoke? A: Well... It's hard to quit. B: But you say to yourself that it is bad for health. Maybe it's worth trying an anti-smoking program? A: Maybe you're right. In fact, I've already thought about it... B: Speaking of quitting smoking, maybe you would like to participate in the "Clean Air" campaign? \\
\midrule

\textit{18. Valence framing}\newline
Emphasizing what a person can lose if they don't do something is more effective than talking about what they will gain if they do it. 
&
"Studies show that women who do not self-examine their breasts are less likely to detect a tumor in the early, treatable phase of the disease." \\
\midrule

\textit{19. The pique technique}\newline
Formulating a message in an unusual way to arouse the interest of the recipient, increasing the likelihood of its acceptance.
&
Scheduling a meeting at 4:55 p.m. instead of 5:00 p.m. \\
\midrule

\textit{20. The only request}\newline
Emphasizing that the request is one-time and does not entail further obligations.
&
"Hello, I'm collecting money for a local children's hospice, we are trying to raise money for its better functioning, will you join us and make a donation? That is the only request I have." \\
\midrule

\multicolumn{2}{l}{\textbf{D. Appeal to social consensus and group norms \textit{(Social norms)}}} \\
\midrule

\textit{21. Metacommunication bind}\newline
Formulating a request for an explanation of the interlocutor's refusal to do us a favor, which is so problematic that she is prompted to comply with our request.
&
You turn to a colleague, "Hey, I need you to look at my results and give me some tips." Your colleague refuses: "I am sorry, but I have too much work", you say: "I understand, but could you tell me why you cannot help me? That's really important to me." \\
\midrule

\textit{22. “Everyone knows it” }\newline 
Appeal to the opinion or behavior of the majority.
&
Club owners create artificial queues outside to suggest high interest and high quality of the venue, which attracts more customers. \\
\midrule

\textit{23. The “We” rule}\newline
Identifying with the characteristics or experiences of a given group in order to increase the interlocutor's propensity to comply with a request.
&
A colleague turns to a coworker: "Everyone in our team is helping with this project, can you join?" \\
\midrule

\textit{24. “That’s how we do it here”}\newline
Drawing attention to an existing social norm and reminding of its meaning.
&
In a housing estate, residents are informed: "In our community, we segregate waste because it is customary for us to do so," which increases the commitment to recycling. \\
\midrule

\textit{25. We are exceptional}\newline
Appeal to the norms of the group to which the recipient belongs, particularly emphasizing its uniqueness.
&
Hotel guests were informed that 75\% of people using their specific room (e.g. No. 215) decided to use the towel again. Appealing to the norm in a small, specific group proved to be more effective than general calls to go green. \\
\midrule

\multicolumn{2}{l}{\textbf{E. Appealing to social reciprocity rule} \textit{\textbf{(Reciprocity})}} \\
\midrule

\textit{26. Birthday effect}\newline
Making requests to a person who has experienced many pleasant gestures during the day.
&
Adam was named employee of the month and receives congratulations throughout the day. At the end of the working day, a colleague asks him for help with one task with which he has a problem. \\
\midrule

\textit{27. Gratitude}\newline
Showing gratitude for a favor done, increasing further involvement.
&
Thanking for work-related activities makes the person think about them more often, see their meaning, and effects more often. \\
\midrule

\textit{28. Give to take}\newline
Doing a small favor to expect future help.
&
Inviting someone to lunch, and then asking for help or a replacement at work. \\
\midrule

\textit{29. Indirect reciprocity}\newline
Taking advantage of someone who has just helped another person, makes them more likely to help us.
&
A driver in a traffic jam willingly lets the car in front of him if he was previously let in by another driver earlier. \\
\midrule

\textit{30. We’ve already given}\newline
Helping someone important to the manipulated person to gain commitment.
&
Person A gives valuable advice to Person B. Person B can't repay A directly, but sees Person C in need of similar help and helps them instead. \\
\midrule

\textit{31. Door-in-the-face}\newline
Presenting a difficult request first, then follow it with the actual, smaller request.
&
A student asks the teacher to release him from his homework completely. After refusing, he asks for an extension of the deadline, which is accepted. \\
\midrule

\multicolumn{2}{l}{\textbf{F. Appealing to emotions \textit{(Emotions)}}} \\
\midrule
\textit{32. Emotional see-saw} \newline Inducing a sudden change of emotions in the interlocutor – from positive to negative or vice versa; putting her in a state of emotional disorientation, making him more susceptible to influence.
&
A teacher tells a student that she failed an important exam (negative emotions), but then adds that the grade was mistaken, and in fact she passed (positive emotions). Then the teacher asks the student: "Can you help me organize the papers? This will help to complete their assessment faster."
\\ \midrule

\textit{33. Fear and anxiety} \newline Inducing a sense of moderately intense anxiety or fear.
&
"If you do not acquire life insurance, your family will be left without financial support in the event of an accident."
\\ \midrule

\textit{34. Anticipatory regret} \newline Inducing in the interlocutor a sense of regret that may occur in the future due to acting or omitting to act now.
&
If you do not start taking care of your health now, then in a few years, when health problems appear, you will regret that you did nothing about it.
\\ \midrule

\textit{35. Take advantage of a good mood} \newline Inducing a positive emotional state in the recipient.
&
A salesperson first tells a funny story or tries to entertain the customer, and then offers to buy the product using their positive emotions.
\\ \midrule

\textit{36. Take advantage of a bad mood} \newline Taking advantage of the interlocutor's existing negative mood to increase compliance.
&
A partner is irritated after an argument with someone else. You ask for a small favor, such as throwing out the garbage, saying that it will take him away from his worries.
\\ \midrule

\textit{37. Physiological arousal} \newline Inducing increased physiological arousal in a person (e.g., accelerated heartbeat).
&
An imaginative depiction of an event, e.g., driving a high-speed sports car.
\\ \midrule

\textit{38. Guilt} \newline Inducing a person to feel guilty in order to increase the interlocutor's propensity to do a favor or fulfill a request as a way to reduce guilt (as a negative emotion).
&
You left me alone in this difficult situation and I was counting on your support and help. Please help me with this task.
\\ \midrule

\textit{39. Shame} \newline Inducing a sense of shame in the interlocutor to increase the interlocutor's propensity to do a favor or fulfill a request as a way to alleviate feelings of shame (as a negative emotion).
&
Your work results cast a shadow over the image of the team. I ask that you complete the next team task on your own.
\\ \midrule

\textit{40. Embarrassment} \newline  Making the interlocutor feel embarrassed, which makes them more likely to agree to the request to improve their image and feel better.
&
I know this may be inconvenient for you, but I really need your help in selecting people from our department to be fired.
\\ \midrule

\textit{41. Show your disappointment} \newline Showing disappointment in the interlocutor's behavior in order to get him to comply with a request, which can improve the mood of both parties.
&
I could always count on you, and now I feel a little disappointed that you don't have time to help me. Can I ask you for support in this task?
\\ \midrule

\textit{42. Positive cognitive state} \newline Arousing a state of intrigue or curiosity in an interlocutor through a trick or riddle that he or she is unlikely to solve. As a result, the interlocutor is more likely to comply with requests when experiencing a mixture of curiosity, surprise, and frustration.
&
"I wonder if you can answer the question my professor once asked me." In a situation where the interlocutor does not find a solution, you suggest "I have an answer for you. In the next step: I would like to ask you to do a little thing for me."
\\ \midrule

\textit{43. Humor} \newline Inducing submission in an individual by 1) weaving a humorous element into the statement OR 2) humorous formulation of a request. This weakens the critical analysis of the content of the message and makes it easier for consent to do a favor.
&
"Hey, I have the impression that this floor is trying to say something to me... But I don't understand the crumb language. Maybe you could help her express herself with a mop?"
\\ \midrule

\textit{44. Foot-in-the-mouth } \newline Arousing the desire to help/do a favor by illustrating the contrast of the recipient's good situation to the difficult situation of people in need of help (e.g., the homeless, the starving, or the terminally ill).
&
A: How do you feel? B: Thank you, good. A: Great! However, not everyone is so lucky! Children in Africa are starving and get sick with deadly diseases. You can support their fate.
\\ \midrule

\textit{45. The power of word "love" } \newline Evoking associations in the interlocutor with the feeling of love or strong positive bond.
&
Requesting a donation for a can marked with the word "love" often results in people tossing money into it.
\\ \midrule
\textit{46. Cognitive exhaustion} \newline Exploiting a person's physical, emotional, or mental exhaustion (or inducing exhaustion) to make requests, which increases the likelihood that the requests will be fulfilled.
&
A: "Could you help me with something small? It is really just a moment." B: "What's the matter?" A: "Great! I need you to fill out this short survey, it is just 5 questions." B: (hesitantly) "Okay, so be it." (B completes the survey, it takes him longer than he expected.) A: "Thank you! And now for the last request – would you please join our list of participants? It is not a big deal, just indicate how many times a month you would like to help with such projects." B: (tired of previous activity) "Phew... Okay, type me in 3 times."
\\ \midrule

\multicolumn{2}{l}{\textbf{G. Appeal to sympathy, liking, connections\textit{ (Liking)}}} \\ \midrule

\textit{47. Liking} \newline Taking advantage of the affection that the recipient has for us to persuade them to comply with our request.
&
People are more likely to buy a product if it is advertised by someone they like.
\\ \midrule

\textit{48. Similarity} \newline The use of commonalities or similarities between the manipulative person and the person to whom the request is directed.
&
"I love playing computer games too! I saw that you have a new game that I wanted to try. Why don't you lend me this game?"
\\ \midrule

\textit{49. Flattery} \newline Providing positive, often exaggerated compliments to the interlocutor to arouse sympathy, favor, or gratitude.
&
"I'm really impressed by the way you manage this project. You have amazing organizational skills, you can always deal with difficult situations. Maybe you can help me with this task?"
\\ \midrule

\multicolumn{2}{l}{\textbf{H. Appeal to authority \textit{(Authority)}}} \\ \midrule

\textit{50. Authority of person or science } \newline The use of prestige, position, or knowledge of authority figures to persuade an interlocutor to accept a particular position or argument.
&
Scientists confirm that the greenhouse effect is a serious threat to life on Earth.
\\ \midrule

\multicolumn{2}{l}{\textbf{I. Appeal to consistency in views and/or behavior \textit{(Consistency)}}} \\ \midrule

\textit{51. That’s not all} \newline Gradual disclosure of elements (benefits) of an offer/proposition in order to increase its attractiveness for the recipient.
&
Our offer is PLN 100 per product, but that is not all! You will also receive free shipping and an additional gadget!
\\ \midrule

\textit{52. Default settings} \newline The use of the human tendency to avoid change and stay with the status quo, especially when taking action involves effort or risk.
&
Insurers often renew policies automatically, and customers who would have to terminate the contract before the due date remain with their current insurer due to inactivity.
\\ \midrule

\textit{53. Inducing involvement} \newline Causing an interlocutor to declare something publicly; to create a sense of public declaration.
&
In a store: "Would you just like to try on this jacket? You don't have to buy right away." Once you have tried it on and decided that it fits, you feel more pressure to buy it.
\\ \midrule

\textit{54. Low ball} \newline Presenting an attractive offer to the interlocutor, which, if accepted, is changed to a less favorable one.
&
A colleague asks for a short help with a project, claiming that it will take 5 minutes. Once you agree, it turns out that the work is more time-consuming, but you feel obliged to help you to the end.
\\ \midrule

55. \textit{Foot-in-the-door} \newline Obtaining permission for the interlocutor to fulfill an easy request and then presenting him with a more demanding request.
&
A neighbor first asks for a small favor, e.g. watering flowers in his absence (small request). After some time, he asks for a greater favor, such as taking care of his animal.
\\ \midrule

\textit{56. Four walls } \newline Induce the interlocutor to make such statements that he falls into the trap of consequences.
&
"Anita, you care most about getting promoted, right?' – yes, "so you certainly want to show how competent you are in analyzing market data" – yes, "you will certainly agree to perform a new task that requires such skills" – probably Anita will not refuse to accept a new task.
\\ \midrule

\textit{57. Even a penny or moment will help} \newline Persuading the interlocutor to commit even a minimal amount of a resource, e.g. time, money, which will result in the achievement of a larger goal.
&
"Literally, a zloty is enough – every penny matters and brings us closer to our goal. Even such a small amount can help provide a meal for a person in need."
\\ \midrule

\textit{58. Make your commitments public} \newline Making your commitment public is more likely to be fulfilled.
&
When someone publicly announces that they are going to exercise every day for a month (e.g. on social media), they feel more pressure to keep their promise not to come across as someone who doesn't keep their word.
\\ \midrule
\caption{Complete SITT technique definitions for each category, with examples.}
\end{longtable}

\clearpage

\section{Prompts}

\subsection{Prompt used for translation}
\label{Prompt:translate}

\begin{tcolorbox}[colback=maincolor!10!white, colframe=maincolor, title=\textbf{English (original)}, width=\textwidth]

Translate the following dialog to the Polish language. The dialogue may contain examples of persuasion or manipulation, and they might be subtle. While translating, change the text so that they are much more prominent, but they also have to sound more natural after the change. Try to only use words an average Polish person would use, and avoid ugly literal translations. Make other changes to make the text sound more natural or to enhance the manipulation - it is not important to translate it one to one. First, think about where the manipulation is, how to amplify it, and then return the dialog without any additional content, preceded by "TRANSLATED DIALOGUE:"\\

Dialogue: \textit{<dialogue>}
\end{tcolorbox}

\subsection{Prompt used to enhance the data}
\label{Prompt:enhance}

\begin{tcolorbox}[colback=maincolor!10!white, colframe=maincolor, title=\textbf{Polish (original)}, width=\textwidth]
Zostanie ci przedstawiony dialog zawierający próbę manipulacji lub perswazji. Dialog został przetłumaczony maszynowo i może zawierać niezręczne, nienaturalne sformułowania, lub błędy. Twoim zadaniem jest poprawić je i sprawić, żeby całość brzmiała dużo bardziej naturalnie. Możesz dokonać znacznych zmian, ale musisz pozostawić manipulację. Zanim zaczniesz pisać, napisz tok rozumowania, w którym przeanalizujesz, jakie błędy i nienaturalne sformułowania widzisz, oraz gdzie jest manipulacja, a także jak możesz je poprawić. Następnie dopiero zwróć dialog, poprzedzony słowem "DIALOG:". \\

Dialog: \textit{<dialogue>}
\end{tcolorbox}

\begin{tcolorbox}[colback=maincolor!10!white, colframe=maincolor, title=\textbf{English (translated)}, width=\textwidth]
You will be presented with a dialogue containing an attempt at manipulation or persuasion. The dialogue was machine-translated and may include awkward, unnatural phrasing or errors. Your task is to correct these and make the entire exchange sound much more natural. You may make significant changes, but you must preserve the manipulation. Before you start writing, provide a reasoning process in which you analyze the errors and unnatural expressions you notice, as well as where the manipulation occurs, and how you can improve the dialogue.\\

Only then should you return the dialogue, preceded by the word 'DIALOG:'.\\

Dialogue: \textit{<dialogue>}
\end{tcolorbox}

\subsection{Prompt used to generate the data}
\label{Prompt:generation}

\begin{tcolorbox}[colback=maincolor!10!white, colframe=maincolor, title=\textbf{Polish (original)}, width=\textwidth]
Hej, proszę Cię o przygotowanie trzech nowych przykładów zastosowania techniki manipulacyjnej, które nie powielają moich wcześniejszych propozycji. Preferuję, aby każdy przykład był przedstawiony jako pojedyncza wypowiedź. Poprzedź go wyrażnym zwrotem "Kontekst:" Postaraj się, aby zapewnić różnorodność w przykładach nie tylko tematów, ale również form. Niech przykład będzie jak z filmu albo życia, a nie z podręcznika. 

\end{tcolorbox}
\rightline{\textit{Continued on the next page}}

\begin{tcolorbox}[colback=maincolor!10!white, colframe=maincolor, title=\textbf{Polish (original)}, width=\textwidth]
Urealnij je maksymalnie. Mogą być nieco ukryte i mgliste tak, jak w rzeczywistości. Niech język też będzie możliwie ludzki, mniej dokładny. Przykłady nie powinny być jak z podręcznika, a bardziej jak z życia lub filmu. Niech to będzie jak najbardziej PRAWDZIWE. Przykłady przez Ciebie podane mogą być bardziej rozbudowane. W przypadku użycia dialogów nie bój się sekwencji dłuższych niż 3-4 wypowiedzi. Dialog początkowo może wyglądać na dość normalny i naturalny, lecz w jego trakcie (np na końcu) ma się pojawić manipulacja. Niech za każdym razem osoby nazywają się: "Osoba A", "Osoba B" itd. Niech postawa i język postaci w dialogach będzie adekwatna do roli czy wieku (niech np. dziecko korzysta z języka odpowiedniego dla jego wieku). Niech przykłady będą jak najbardziej naturalne. W przypadku użycia dialogów nie bój się sekwencji dłuższych niż 3-4 wypowiedzi. Jeśli będą potrzebne, niech podane konteksty mają sens względem podanej techniki. Przemyśl dobrze, czy kontekst sytuacji dobrze wskazuje na sens użycia tej techniki manipulacyjnej. Podaj również, o ile to możliwe motyw manipulatora, poprzedzając to zwrotem "Motyw:". Nie dodawaj żadnych wytłumaczeń w dialogach. Niech wyglądają możliwie naturalnie. Nie dodawaj zwrotów typu "kolega/koleżanka", bo nikt tak nie mówi. Niech forma każdego z przykładów będzie możliwie różna. Staraj się nie powielać kalk gramatycznych czy lingwistycznych, niech teksty będą zupełnie inne.\\

W pierwszym kroku, zamiast podawać konkretną odpowiedź, przeprowadź tok rozumowania, opisz w jaki sposób rozwiązać powyższe zadanie, opisz swoją własną definicję zadania i to, jak je rozumiesz. Nie ograniczaj się co do szerokości swoich rozmyślań. Spróbuj wyróżnić na tym etapie różne schematy wykorzystania danej techniki. Możesz na tym etapie zarysować wstępnie schematy zastosowań. Na ich podstawie przygotuj prototypy przykładów, które później rozbudujesz. Kiedy już skończysz, wyraźnie zaznacz rozpoczęcie odpowiedzi poprzez utworzenie sekcji "Przykłady:"\\

Nazwa techniki: <\textit{technique\_name}>\\

Definicja techniki: <\textit{definition}>\\

Przykłady, które już opisałem: <\textit{samples}>\\
\end{tcolorbox}

\begin{tcolorbox}[colback=maincolor!10!white, colframe=maincolor, title=\textbf{English (translated)}, width=\textwidth]

"Hi, I’d like you to prepare three new examples of the use of the manipulation technique, which do not repeat any of my earlier examples. I prefer that each example be presented as a single utterance. Precede each one with a clear label 'Context:'. Try to ensure diversity in the examples not only in terms of topic, but also in form. Let the examples feel like they’re from a movie or real life, not a textbook. Make them as realistic as possible. They can be somewhat hidden and vague, just like in real life. The language should also be as natural and human as possible—less precise. The examples should not read like they’re from a textbook, but more like they’re from everyday life or film. Make them as REAL as possible. The examples you provide may be more developed. If you use dialogues, don’t be afraid of sequences longer than 3–4 lines. The dialogue may initially seem normal and natural, but as it progresses (e.g., toward the end), the manipulation should emerge. Use the names “Person A,” “Person B,” etc., every time. The attitude and language of the characters in the dialogues should match their role or age (e.g., a child should speak like a child). Keep the examples as natural as possible. If you use dialogues, don’t be afraid of sequences longer than 3–4 lines.
\end{tcolorbox}
\rightline{\textit{Continued on the next page}}

\begin{tcolorbox}[colback=maincolor!10!white, colframe=maincolor, title=\textbf{English (translated)}, width=\textwidth]
If needed, the provided contexts should make sense in relation to the manipulation technique. Think carefully about whether the situation context clearly supports the use of this manipulation technique. Also provide, if possible, the manipulator’s motive, prefaced by the label 'Motive:'. Do not add any explanations in the dialogues. They should sound as natural as possible. Do not include expressions like “friend” or “buddy” unless that’s how people actually speak in that context. Make each example as different in form as possible. Try not to repeat grammatical or linguistic patterns—each text should be completely different.\\
 
In the first step, instead of giving specific examples, go through your reasoning process. Describe how you would approach solving the task above. Provide your own definition of the task and how you understand it. Don’t limit the scope of your thinking. Try to distinguish various patterns of how this technique could be used. At this stage, you may sketch out preliminary usage patterns. Based on these, prepare prototype examples that you will later expand. Once you're finished, clearly indicate the beginning of your answer by creating a section titled 'Examples:'"\\

Technique name: <\textit{technique\_name}>\\

Technique definition: <\textit{definition}>\\

Examples I have already described: <\textit{samples}>\\
\end{tcolorbox}

\subsection{Prompt for category class first distribution}
\label{Prompt:Category_Label}

\begin{tcolorbox}[colback=maincolor!10!white, colframe=maincolor, title=\textbf{Polish (original)}, width=\textwidth]
Przedstawiony Ci zostanie tekst o potencjalnym charakterze manipulacyjnym wraz z listą potencjalnych kategorii manipulacyjnych wraz z przykładami i definicjami. Twoim zadaniem jest przypisanie kategorii technik manipulacyjnych do zadanego tekstu.
Z racji, że w tekście może być zastosowana więcej niż jedna kategoria technik, możesz podać kilka kategorii. Wynik podaj w postaci listy numerów przypisanych do konkretnych kategorii.
Przed podaniem listy podaj tok rozumowania. Pod jego koniec podaj konkretne fragmenty, które wskazują zastosowaną technikę, wraz z wytłumaczeniem, dlaczego dany fragment pasuje do danej kategorii. Poprzedź ten fragment zwrotem: "Znalezione manipulacje:".
Na końcu podaj listę kategorii (tylko numerki). Listę podaj, poprzedzając zwrotem: "Lista:". Jeśli w tekście nie ma techniki manipulacyjnej, nie umieszczaj nic w sekcji "Znalezione manipulacje:", a jako listę wpisz '[0]'.\\

Lista technik:\\

1. Odwoływanie się do pozytywnego lub negatywnego wizerunku odbiorcy
Polega na wykorzystaniu istniejących lub oczekiwanych cech odbiorcy w celu wywołania określonych reakcji. W tym procesie nadawca odwołuje się do określonych cech, wartości, poglądów, itd. odbiorcy w celu nakłonienia go do zachowań zgodnych z własnym wizerunkiem.
Przykład: “Z pewnością wiesz, jako ekspert bezpieczeństwa, który zna i stosuje reguły bezpieczeństwa w sieci, że nie należy stosować prostych haseł.”\\

\end{tcolorbox}
\rightline{\textit{Continued on the next page}}

\begin{tcolorbox}[colback=maincolor!10!white, colframe=maincolor, title=\textbf{Polish (original)}, width=\textwidth]
2. Manipulowanie kontekstem
Techniki manipulowania kontekstem polegają na modyfikowaniu lub wykorzystaniu elementów otoczenia, sytuacji, czasu, miejsca lub przestrzeni w taki sposób, aby wpłynąć na postrzeganie i decyzje odbiorców. Są one szczególnie skuteczne, ponieważ wpływają na interpretację i emocjonalny odbiór informacji, często nieświadomie kształtując decyzje i zachowania. Nie zmienia się informacja tylko kontekst, w którym jest przedstawiana.
Przykład: Sprzedawca rozmawia z klientem, pytając o rzeczy, z którymi ten łatwo się zgadza, np.: „Czy zdrowie Twojej rodziny jest dla Ciebie ważne?”. Po kilku odpowiedziach „tak” klient trudniej odmawia zakupowi ubezpieczenia zdrowotnego.\\

3. Manipulowanie informacją i/lub argumentacją
Techniki w tej kategorii bazują na przedstawieniu informacji w sposób tendencyjny, sugestywny, selektywny, dwuznaczny lub upraszczający. W zależności od kontekstu informacja i argumenty pochodzące od nadawcy lub odbiorcy zostają zniekształcone (np. wzmocnione, osłabione, lub uwaga odbiorcy zostaje przekierowana).
Prykład: Ktoś odmawia pomocy, tłumacząc, że nie ma czasu. Możemy zaproponować skrócenie czasu potrzebnego na spełnienie prośby, np. do kilku minut\\

4. Odwoływanie się do większości i norm grupowych
Techniki manipulacji opierające się na odwoływaniu do norm społecznych wykorzystują skłonność ludzi do naśladowania zachowań większości lub przestrzegania norm akceptowanych w danej grupie. Normy społeczne działają jako mechanizm regulujący zachowania, a ich zastosowanie w komunikacji może skłaniać ludzi do podejmowania działań zgodnych z oczekiwaniami grupy.
Przykład: „80\% mieszkańców Twojego osiedla oszczędza energię, dlatego prosimy Cię o włączenie się do naszej akcji.”\\

5. Odwoływanie się do reguły wzajemności
Techniki manipulacji oparte na regule wzajemności bazują na zasadzie, zgodnie z którą ludzie czują się zobowiązani do odwzajemnienia przysług, gestów lub korzyści, które otrzymali od innych. Ta zasada jest głęboko zakorzeniona w normach społecznych, ponieważ odwzajemnianie jest kluczowym mechanizmem regulującym wymianę społeczną i budującym relacje w grupach.
Przykład: “Proszę, oto drobny upominek od naszej restauracji – breloczek, który przypomni Panu o naszej ofercie. Czy mogę zaproponować nasze specjalne menu na dziś?” Mechanizm: Klient, otrzymując prezent, czuje się zobowiązany do odwzajemnienia gestu, np. zamówienia droższych dań.\\

6. Odwoływanie się do emocji
Odwoływanie się do emocji, to taktyka psychologiczna, w której jednostka wykorzystuje bodźce emocjonalne - takie jak strach, złość, litość lub radość - aby wpłynąć na postawy, decyzje lub zachowania innych, jednocześnie obniżając racjonalną analizę lub krytyczne myślenie. Metoda ta jest często wykorzystywana do przekonywania lub manipulowania poprzez wywoływanie silnych uczuć zamiast przedstawiania logicznych argumentów lub faktycznych dowodów.
Przykład 1: wywołanie lęku wśród potencjalnych wyborców
Polityk twierdzi: „Jeśli nie zagłosujesz na naszą partię, kraj pogrąży się w chaosie, a bezpieczeństwo twojej rodziny będzie zagrożone”.\\

7. Techniki oparte na regule lubienia, sympatii, więzi
Te techniki opierają się one na tendencji ludzi do bycia bardziej przekonanymi przez tych, których lubią, znają lub z którymi mają częsty kontakt. Taktyki te omijają krytyczne myślenie, wykorzystując pozytywne uczucia lub komfort związany ze źródłem lub przekazem.
Przykład 1: Sprzedawca buduje relacje poprzez komplementowanie klienta, rzekome dzielenie wspólnych zainteresowań oraz bycie ciepłym i przyjaznym.
\end{tcolorbox}
\rightline{\textit{Continued on the next page}}

\begin{tcolorbox}[colback=maincolor!10!white, colframe=maincolor, title=\textbf{Polish (original)}, width=\textwidth]
8. Odwoływanie się do autorytetu/atrybutów autorytetu
Techniki oparte  na autorytecie polegają na wykorzystaniu władzy, pozycji lub eksperckiej wiedzy danej osoby lub instytucji do wpływania na zachowanie, decyzje czy przekonania innych ludzi.
Przykład: "Znany ekspert w dziedzinie biologii, twierdzi, że zmiany klimatyczne są realnym zagrożeniem dla naszej planety, dlatego powinniśmy podjąć działania na rzecz ochrony środowiska."\\

9.  Odwoływanie się do konsekwencji w poglądach i zachowaniach
Techniki manipulacji opierające się na regule zaangażowania i konsekwencji polegają na uzyskaniu wstępnego zaangażowania osoby w określone działanie, co prowadzi do większej gotowości do spełniania kolejnych, bardziej wymagających próśb.
Przykład:  „Czy mógłby Pan poświęcić minutę na podpisanie naszej petycji? A skoro już Pan ją podpisał, czy rozważyłby Pan również udział w naszej kampanii informacyjnej?”\\

Tekst do oceny: \textit{<text>}
\end{tcolorbox}

\begin{tcolorbox}[colback=maincolor!10!white, colframe=maincolor, title=\textbf{English (translated)}, width=\textwidth]

You will be presented with a text that may have manipulative characteristics, along with a list of potential manipulation technique categories, including definitions and examples. Your task is to assign the relevant manipulation categories to the given text.
Since the text may include more than one manipulation technique, you may provide multiple categories. Present your answer as a list of numbers corresponding to the relevant categories.
Before providing the list, explain your reasoning. At the end of your reasoning, indicate the specific fragments that show the use of a manipulation technique, and explain why each fragment fits the selected category. Begin this section with the phrase: “Detected manipulations:”.
At the end, provide the list of categories (just the numbers), preceded by the word: “List:”. If there are no manipulation techniques in the text, leave the “Detected manipulations:” section empty and write '[0]' as the list.\\

List of techniques:\\

1. Appealing to the recipient’s positive or negative self-image
This involves using actual or expected traits of the recipient to provoke specific reactions. The sender refers to certain traits, values, or beliefs of the recipient to influence behavior consistent with the desired self-image.
Example: “As a cybersecurity expert who knows and applies safety rules, you surely understand that using simple passwords is unwise.”\\

2. Manipulating the context
These techniques involve modifying or using elements of the environment, situation, time, location, or space to influence how recipients perceive and respond. These are effective because they shape emotional interpretation, often subconsciously. The information itself doesn’t change — only the context.
Example: A salesperson asks a customer questions that are easy to agree with, like: “Is your family’s health important to you?” After several “yes” answers, it becomes harder to decline a health insurance offer.\\

3. Manipulating information and/or argumentation
These techniques rely on presenting information in a biased, suggestive, selective, ambiguous, or oversimplified way. Depending on context, the sender's or recipient's information or arguments may be distorted (e.g., exaggerated, weakened, or redirected).
Example: Someone refuses to help, saying they don’t have time. You then propose a version of the request that takes only a few minutes.
\end{tcolorbox}
\rightline{\textit{Continued on the next page}}

\begin{tcolorbox}[colback=maincolor!10!white, colframe=maincolor, title=\textbf{English (translated)}, width=\textwidth]
4. Appealing to the majority or group norms
These techniques use social norms and the human tendency to conform to group behavior. Norms regulate behavior, and referring to them can influence people to act in line with group expectations.
Example: “80\% of your neighbors are saving energy — join them in our campaign.”\\

5. Appealing to the reciprocity rule
Based on the social rule that people feel obliged to return favors or gestures. This is a key mechanism for social exchange and building relationships.
Example: “Here’s a small gift from our restaurant – a keychain to remind you of our offer. May I suggest today’s special?” Mechanism: The customer feels compelled to reciprocate, e.g., by ordering more expensive dishes.\\

6. Appealing to emotions
This psychological tactic uses emotional triggers — such as fear, anger, pity, or joy — to influence attitudes or behaviors, while lowering rational analysis. It often replaces facts or logic with emotional impact.
Example: A politician says: “If you don’t vote for us, the country will fall into chaos and your family’s safety will be at risk.”\\

7. Techniques based on liking, sympathy, or rapport
These techniques exploit our tendency to be more persuaded by people we like, know, or interact with frequently. They bypass critical thinking by leveraging positive feelings or comfort related to the speaker or message.
Example: A salesperson builds rapport by complimenting the client, pretending to share interests, and acting warm and friendly.\\

8. Appealing to authority/attributes of authority
These rely on power, status, or expertise to influence beliefs or actions.
Example: “A renowned biology expert says climate change is a real threat to our planet, so we must take action to protect the environment.”\\

9. Appealing to consistency in beliefs and behavior
These are based on the rule of commitment and consistency — once someone agrees to something small, they’re more likely to agree to bigger requests.
Example: “Could you take a minute to sign our petition? Since you've signed, would you consider joining our awareness campaign too?”\\

Text to analyze:\textit{<text>}
\end{tcolorbox}

\subsection{Prompt translation from Polish}
\label{Prompt:translate_from_polish}

\begin{tcolorbox}[colback=maincolor!10!white, colframe=maincolor, title=\textbf{English (original)}, width=\textwidth]
Translate the following text to english. Try to be as accurate as possible.\\

Text: \textit{<text>}\\

English text:
\end{tcolorbox}

\subsection{Prompt for categories classification}
\label{Prompt:categories_classification}

\begin{tcolorbox}[colback=maincolor!10!white, colframe=maincolor, title=\textbf{Polish (original)}, width=\textwidth]
Instrukcja

Zostanie Ci przedstawiony tekst, który może (ale nie musi) zawierać techniki wpływu społecznego. 
\end{tcolorbox}
\rightline{\textit{Continued on the next page}}

\begin{tcolorbox}[colback=maincolor!10!white, colframe=maincolor, title=\textbf{Polish (original)}, width=\textwidth]
Twoim zadaniem jest ocenić, czy w tekście występuje którykolwiek z poniższych rodzajów wpływu społecznego. Jeśli tak, wskaż odpowiednie numery kategorii. Jeśli tekst nie zawiera żadnej z wymienionych technik, wpisz [0]. Nie szukaj na siłę – możliwe, że tekst nie zawiera wpływu społecznego.
Możliwe klasy (numery) wraz z definicjami i przykładami:\\

1. Odwoływanie się do pozytywnego/negatywnego wizerunku odbiorcy\\
Definicja: Zespół technik wpływu społecznego, które polegają na odwoływaniu się do samooceny odbiorcy – jego poczucia tożsamości, godności, moralności lub społecznego wizerunku – w celu skłonienia go do określonego zachowania. W przekazie wykorzystuje się zarówno pozytywne etykietowanie (np. „jesteś odpowiedzialną osobą”), jak i negatywne (np. „tylko ignorant by tego nie zrobił”), aby wywołać presję do działania zgodnego z narzuconą etykietą.\\
Przykłady:\\
    a. “Z pewnością wiesz, jako ekspert bezpieczeństwa, który zna i stosuje reguły bezpieczeństwa w sieci, że nie należy stosować prostych haseł.”\\
    b. “Tę ofertę przygotowaliśmy wyłącznie dla Pana: nikt inny nie może z niej skorzystać.”\\
    c. “Prawdopodobnie odmówisz, ale ciekaw jestem, czy byłbyś skłonny nam pomóc, ofiarowując datek pieniężny.”\\

2. Modyfikowanie kontekstu\\
Definicja: Zespół technik wpływu społecznego polegającego na modyfikowaniu lub wykorzystaniu elementów otoczenia, sytuacji, czasu, miejsca lub przestrzeni w taki sposób, aby wpłynąć na postrzeganie i decyzje odbiorców. Są one szczególnie skuteczne, ponieważ wpływają na interpretację i emocjonalny odbiór informacji, często nieświadomie kształtując decyzje i zachowania. Nie zmienia się informacja tylko kontekst, w którym jest przedstawiana.\\
Przykłady:\\
	a. Sprzedawca rozmawia z klientem, pytając o rzeczy, z którymi ten łatwo się zgadza, np.: „Czy zdrowie Twojej rodziny jest dla Ciebie ważne?”. Po kilku odpowiedziach „tak” klient trudniej odmawia zakupowi ubezpieczenia zdrowotnego.\\
	b. Produkty premium są umieszczone na w	ysokości wzroku klienta, podczas gdy tańsze opcje znajdują się na dolnych półkach, co podświadomie zachęca do wyboru droższych produktów.\\
	c. Tylko dziś możesz kupić bilet na koncert ze zniżką 50\%!” – ograniczenie czasowe tworzy presję, by podjąć decyzję natychmiast.\\

3. Tendencyjne przedstawianie informacji i/lub argumentów
Definicja: Techniki w tej kategorii bazują na przedstawieniu informacji w sposób tendencyjny, sugestywny, selektywny, dwuznaczny lub upraszczający. W zależności od kontekstu informacja i argumenty pochodzące od nadawcy lub odbiorcy zostają zniekształcone (np. wzmocnione, osłabione, lub uwaga odbiorcy zostaje przekierowana). W ten sposób obraz danej rzeczywistości zostaje zafałszowany, co może wpłynąć na postrzeganie, opinie i decyzje odbiorcy.\\
Przykłady:\\
	a. A: Czy przesuniemy termin oddania projektu o jeden dzień?\\
	B: To może być problematyczne.\\
	A: Dodatkowy dzień umożliwi zebranie wszystkich niezbędnych informacji.\\
	Przykład: Dietetyk mówi osobie odchudzającej się: „Wiem, jak trudno zrezygnować z czekolady, bo jest pyszna i poprawia nastrój, ale aby poprawić swoje zdrowie i uniknąć cukrzycy, warto wprowadzić zmiany w diecie.”

\end{tcolorbox}
\rightline{\textit{Continued on the next page}}

\begin{tcolorbox}[colback=maincolor!10!white, colframe=maincolor, title=\textbf{Polish (original)}, width=\textwidth]
	b. A: Wiem, że palenie jest szkodliwe, każdy to wie.\\
	B: To czemu nadal palisz?\\
	A: No… to trudne do rzucenia.\\
	B: Ale sam mówisz, że to złe dla zdrowia. Może warto spróbować jakiegoś programu antynikotynowego?\\
	A: Może masz rację. W sumie już o tym myślałem\dots\\
	B: Skoro już jesteśmy przy rzuceniu palenia, może zechciałbyś zaangażować się w akcję „Czyste powietrze”?\\
	c. „Badania pokazują, że kobiety, które nie przeprowadzają samobadania piersi, mają mniejszą szansę na wykrycie guza we wczesnej, poddającej się leczeniu, fazie choroby"\\
    
4. Odwoływanie się do większości i/lub norm grupowych\\
Definicja: Techniki wpływu społecznego opierające się na odwoływaniu do norm społecznych wykorzystujące skłonność ludzi do naśladowania zachowań większości lub przestrzegania norm akceptowanych w danej grupie. Normy społeczne działają jako mechanizm regulujący zachowania, a ich zastosowanie w komunikacji może skłaniać ludzi do podejmowania działań zgodnych z oczekiwaniami grupy.
Przykłady:\\
	a. „80\% mieszkańców Twojego osiedla oszczędza energię, dlatego prosimy Cię o włączenie się do naszej akcji.”\\
	b. „W naszej firmie wszyscy pracownicy segregują odpady -to standard, który wspiera nasze wartości ekologiczne.”\\
	c. „Każdy w naszym biurze wpłacił już datek na pomoc potrzebującym -Twoja wpłata może wiele zmienić!”\\
    
5. Odwoływanie się do społecznej wzajemności\\
Definicja: Zespól technik wpływu społecznego opartych na regule wzajemności bazują na zasadzie, zgodnie z którą ludzie czują się zobowiązani do odwzajemnienia przysług, gestów lub korzyści, które otrzymali od innych. Ta zasada jest głęboko zakorzeniona w normach społecznych, ponieważ odwzajemnianie jest kluczowym mechanizmem regulującym wymianę społeczną i budującym relacje w grupach.\\
Przykłady:\\
	a. “Proszę, oto drobny upominek od naszej restauracji – breloczek, który przypomni Panu o naszej ofercie. Czy mogę zaproponować nasze specjalne menu na dziś?” Mechanizm: Klient, otrzymując prezent, czuje się zobowiązany do odwzajemnienia gestu, np. zamówienia droższych dań.\\
	b. “Zdajemy sobie sprawę, że wpłaty na cele charytatywne to osobista decyzja, ale chcemy podziękować za wcześniejsze wsparcie i zapytać, czy moglibyśmy liczyć na Państwa darowiznę również w tym roku.” Mechanizm: Wysłanie podziękowania za poprzednie darowizny tworzy zobowiązanie do dalszego wspierania akcji.\\
	c. “Otrzymał(a) Pan(i) od nas darmową próbkę nowego kremu. Jak wrażenia? Możemy zaproponować pełnowartościowy produkt w promocyjnej cenie.” Mechanizm: Darmowa próbka tworzy zobowiązanie do zakupu produktu w pełnym wymiarze.\\
    
    6. Odwoływanie się do emocji
Definicja: To kategoria technik wpływu społecznego, która polega na celowym wywoływaniu u odbiorcy określonych stanów emocjonalnych (np. lęku, winy, wzruszenia, dumy, entuzjazmu), aby zwiększyć podatność na sugestię, przekonać do określonego działania lub wpłynąć na decyzje. Emocje te odwracają uwagę od racjonalnej analizy informacji i wzmacniają skłonność do działania zgodnie z intencją nadawcy.\\
Przykład: wywołanie lęku wśród potencjalnych wyborców
\end{tcolorbox}
\rightline{\textit{Continued on the next page}}

\begin{tcolorbox}[colback=maincolor!10!white, colframe=maincolor, title=\textbf{Polish (original)}, width=\textwidth]
Przykład: wywołanie lęku wśród potencjalnych wyborców
Polityk twierdzi: „Jeśli nie zagłosujesz na naszą partię, kraj pogrąży się w chaosie, a bezpieczeństwo twojej rodziny będzie zagrożone”.
To stwierdzenie manipuluje strachem, aby skłonić ludzi do działania w określony sposób, pomijając racjonalną ocenę polityki.\\

7. Odwoływanie się do sympatii, lubienia lub więzi społecznych\\
Definicja: To kategoria technik perswazyjnych polegająca na wykorzystywaniu emocjonalnej bliskości, sympatii lub poczucia podobieństwa między nadawcą a odbiorcą komunikatu. Celem tych działań jest zwiększenie skuteczności przekazu poprzez budowanie pozytywnego nastawienia, wzbudzenie zaufania lub poczucia wspólnoty. Ludzie częściej ulegają wpływowi osób, które lubią, z którymi się utożsamiają lub które okazują im sympatię.\\
Przykłady:\\
	a. Ludzie są bardziej skłonni kupić produkt, jeśli jest on reklamowany przez kogoś, kogo lubią.\\
	b. „Też uwielbiam grać w gry komputerowe! Widziałem, że masz nową grę, którą chciałem wypróbować. Może pożyczysz mi tę grę?"\\
	c. „Naprawdę imponuje mi sposób, w jaki zarządzasz tym projektem. Masz niesamowite zdolności organizacyjne, zawsze potrafisz poradzić sobie w trudnych sytuacjach. Może pomożesz mi z tym zadaniem?”\\
    
8. Odwoływanie się do autorytetu/atrybutów autorytetu\\
Definicja: Odwoływanie się do opinii, stanowiska lub polecenia osoby postrzeganej jako autorytet w danej dziedzinie (np. ekspert, naukowiec, lider, lekarz), aby zwiększyć wiarygodność komunikatu i skłonić odbiorcę do zaakceptowania określonego stanowiska lub podjęcia działania. Również, wywoływanie posłuszeństwa lub zaufania poprzez eksponowanie zewnętrznych oznak autorytetu, takich jak tytuł naukowy, uniform, stanowisko, instytucja czy sposób wypowiedzi, zamiast faktycznej wiedzy, kompetencji lub argumentów.\\
Przykłady:\\
	a. Naukowcy potwierdzają, że efekt cieplarniany jest poważnym zagrożeniem dla życia na Ziemi.\\
    
9. Odwoływanie się do konsekwencji w poglądach i/lub zachowaniach\\
Definicja: Zespół technik perswazyjnych, które wykorzystują ludzką potrzebę zachowania spójności pomiędzy wcześniejszymi deklaracjami, działaniami a aktualnymi decyzjami. Celem tych technik jest skłonienie odbiorcy do kontynuowania wcześniej rozpoczętego działania, wyrażonego stanowiska lub zaakceptowania bardziej angażujących próśb, bazując na mechanizmach konsekwencji, zaangażowania i niechęci do zmiany zdania lub kierunku działania. Techniki te wzmacniają motywację do działania poprzez wytworzenie presji psychologicznej wynikającej z potrzeby wewnętrznej spójności oraz chęci bycia postrzeganym jako osoba konsekwentna i wiarygodna.\\

Format odpowiedzi:\\
Podaj odpowiedź w formacie: \#Odpowiedź: [x,y,z], gdzie x, y, z to numery z listy kategorii. Liczba kategorii może być różna, także nie przywiązuj się do 3. Jeśli żadna kategoria nie pasuje, wpisz \#Odpowiedź: [0].\\

Jeśli odpowiedź to [0], dopisz też:\\
"\#Co musiałoby się zmienić, aby w tekście wystąpił wpływ społeczny:" i podaj, co należałoby zmienić lub dodać, aby w tekście pojawił się jakiś wpływ społeczny.\\

Tekst: \textit{<text>}
\end{tcolorbox}

\begin{tcolorbox}[colback=maincolor!10!white, colframe=maincolor, title=\textbf{English (translated)}, width=\textwidth]
Instruction\\
You will be presented with a text that may (but does not have to) contain techniques of social influence. Your task is to assess whether the text includes any of the types of social influence listed below. If it does, indicate the appropriate category numbers. If the text does not contain any of the mentioned techniques, enter [0]. Don’t try to force it – it is possible that the text contains no social influence.\\

Possible classes (numbers) with definitions and examples:\\

1. Appeal to a positive or negative image\\
Definition: A set of techniques that refer to the recipient's self-evaluation in terms of their sense of identity, dignity, morality or social image – in order to induce them to behave in a certain way consistent with a positive image (e.g. "you are a responsible person") or contradicting a negative image of a person (e.g. "only an ignorant person would not do it"”).\\
Examples:\\
a. “Surely you know, as a cybersecurity expert who knows and applies internet safety rules, that using simple passwords is not advised.”\\
b. “This offer was prepared exclusively for you: no one else can take advantage of it.”\\
c. “You’ll probably say no, but I’m curious – would you be willing to help us by making a donation?”\\

2. Modifying the context\\
Definition: A set of techniques based on modifying or using elements of context (i.e. situation, time, place or space) in such a way as to influence the perception and decisions of the audience. The information does not change, only the context in which it is presented. They are particularly effective because they impose the reception and interpretation of information.\\
Examples:\\
a. A salesperson talks to a client, asking about things the client easily agrees with, e.g.: “Is your family’s health important to you?” After several “yes” answers, it becomes harder for the client to refuse to buy health insurance.\\
b. Premium products are placed at eye level, while cheaper options are on lower shelves, subconsciously encouraging the selection of more expensive products.\\
c. “Only today you can buy a concert ticket at a 50\% discount!” – the time limit creates pressure to make a decision immediately.\\

3. Biased presentation of information and/or arguments\\
Definition: A group of techniques related to presenting information in a biased, suggestive, selective, ambiguous or simplifying way. The information or argument is distorted (e.g. strengthened, weakened, incomplete, with the subjective sense of the sender). In this way, the image of reality/issue is falsified in order to evoke a specific opinion or decision of the recipient\\
Examples:\\
a. A: Can we postpone the project deadline by one day?\\
B: That might be problematic.\\
A: An extra day would allow us to gather all the necessary information.\\
Another example: A dietitian says to someone trying to lose weight: “I know it’s hard to give up chocolate because it’s delicious and improves your mood, but to improve your health and avoid diabetes, it’s worth changing your diet.”
\end{tcolorbox}
\rightline{\textit{Continued on the next page}}

\begin{tcolorbox}[colback=maincolor!10!white, colframe=maincolor, title=\textbf{English (translated)}, width=\textwidth]
b: A: I know smoking is harmful, everyone does.\\
B: Then why do you still smoke?\\
A: Well… it’s hard to quit.\\
B: But you just said it’s bad for your health. Maybe try a smoking cessation program?\\
A: Maybe you’re right. I’ve actually thought about it\dots\\
B: Since we’re talking about quitting smoking, maybe you’d like to join the “Clean Air” campaign?
c. “Studies show that women who don’t perform breast self-exams have a lower chance of detecting tumors at an early, treatable stage of the disease.”\\

4. Appeal to social consensus and group norms\\
Definition: A set of techniques that induce behaviors based on the tendency to imitate the behavior of most people or to follow commonly accepted social norms.\\
Examples:\\
a. “80\% of residents in your neighborhood save energy, so we ask you to join our initiative.”\\
b. “In our company, all employees sort waste – it’s a standard that supports our ecological values.”\\
c. “Everyone in our office has already donated to help those in need – your donation can make a big difference!”\\

5. Appealing to social reciprocity\\
Definition: A set of techniques based on the principle of reciprocity, which uses a sense of obligation to reciprocate favors, gestures, or benefits they have received from others.\\
Examples:\\
a. “Here’s a small gift from our restaurant – a keychain to remind you of our offer. May I suggest today’s special menu?” Mechanism: The client, receiving a gift, feels obligated to reciprocate, e.g., by ordering more expensive dishes.\\
b. “We understand that donating to charity is a personal decision, but we want to thank you for your previous support and ask if we can count on your donation again this year.” Mechanism: Thanking for previous donations creates an obligation to continue supporting the cause.\\
c. “You’ve received a free sample of our new cream. How do you like it? We can offer the full-size product at a promotional price.” Mechanism: A free sample creates an obligation to buy the full product.\\

6. Appeal to emotions\\
Definition: A set of techniques consisting in deliberately evoking a positive or negative mood or specific emotions in the recipient (such as fear, guilt, emotion, grief, disappointment) in order to convince him or her to a specific argument or action that trigger thinking and behavior, reducing negative feelings and intensifying positive feelings.\\
Example: Inducing fear among potential voters.\\

A politician claims: “If you don’t vote for our party, the country will descend into chaos, and your family’s safety will be at risk.”\\
This statement manipulates fear to drive action, bypassing rational political evaluation.\\

7. Appeal to sympathy, liking, connections\\
Definition: A category of techniques consisting in inducing in the recipient sympathy or a sense of similarity or emotional closeness with the sender in order to persuade the recipient to act with the sender's intentions.\\
Examples:
\end{tcolorbox}
\rightline{\textit{Continued on the next page}}

\begin{tcolorbox}[colback=maincolor!10!white, colframe=maincolor, title=\textbf{English (translated)}, width=\textwidth]
a. People are more likely to buy a product if it is advertised by someone they like. \\
b. “I also love playing computer games! I saw you’ve got that new game I wanted to try. Maybe you could lend it to me?”\\
c. “I’m really impressed with how you manage this project. You have amazing organizational skills and always know how to handle tough situations. Maybe you could help me with this task?”\\

8. Appeal to authority\\
Definition: Techniques that refer to [1] knowledge, social position of individuals and/or institutions, [2] facts, scientific findings or scientific sources, and [3] to apparent attributes of authority (as academic titles, positions, institutions) in order to increase the credibility of arguments and convince the recipient to them.\\
Examples:\\
a. Scientists confirm that the greenhouse effect is a serious threat to life on Earth.\\

9. Appeal to consistency in views and/or behavior\\
Definition: A set of techniques referring to or inducing a natural human need to maintain consistency of one's beliefs or behaviors and the consistency of declarations with actions.\\

Response format:\\
Provide your answer in the format: \#Answer: [x,y,z], where x, y, z are numbers from the category list. The number of categories may vary, so don’t assume it must be three. If none of the categories apply, write \#Answer: [0].\\

If your answer is [0], also write:\\
"\#What would have to change for social influence to occur in the text:"
and indicate what would need to be changed or added for some form of social influence to appear in the text.\\

Text: \textit{<text>}
\end{tcolorbox}

\subsection{Prompt technique classification}
\label{Prompt:technique_classification}

\begin{tcolorbox}[colback=maincolor!10!white, colframe=maincolor, title=\textbf{Polish (original)}, width=\textwidth]
Przestawiony Ci zostanie tekst przedstawiający wpływ społeczny. Twoim zadaniem jest ocena która spośród przedstawionych technik wpływu społecznego znajduje się w tekście.\\

Techniki wpływu społecznego: \textit{<techniques>}\\

Podaj odpowiedź w formacie: \#Odpowiedź: [x,y,z], gdzie x, y, z to numery z listy. Liczba technik może być różna, także nie przywiązuj się do 3. Po podaniu listy podaj wyjaśnienie dlaczego uważasz, że powyższe techniki zostały użyte w podanym tekście.\\

Tekst: \textit{<text>}
\end{tcolorbox}

\begin{tcolorbox}[colback=maincolor!10!white, colframe=maincolor, title=\textbf{English (translated)}, width=\textwidth]
You will be presented with a text demonstrating social influence. Your task is to assess which of the listed social influence techniques are present in the text.
\end{tcolorbox}
\rightline{\textit{Continued on the next page}}

\begin{tcolorbox}[colback=maincolor!10!white, colframe=maincolor, title=\textbf{English (translated)}, width=\textwidth]
Social influence techniques: \textit{<techniques>}\\

Provide your answer in the format: \#Answer: [x, y, z], where x, y, z are the numbers from the list. The number of techniques may vary, so don’t assume there will always be three. After listing them, explain why you believe the selected techniques were used in the provided text.\\

Text: \textit{<text>}
\end{tcolorbox}

\section{Expert verification}
\label{sec:expert-verification}
\begin{table}[h]
\centering
\begin{tabular}{lrrrr}
\toprule
\multirow{2}{*}{\textbf{Annotator ID}} & \multicolumn{4}{c}{\textbf{Number of annotations:}} \\
 & \textbf{Dialogues} & \textbf{Techniques} & \textbf{Verified} & \textbf{Consistent with definition} \\
 \midrule
1            & 215& 128& 15&  15 (100\%)\\
2            & 174& 112& 16&  16 (100\%)\\
3            & 174& 184& 38&  34 (89\%)\\
4            & 252& 216& 27&  25 (93\%)\\
5            & 215& 66& 31&  28 (89\%)\\
6            & 215& 43& 39&  33 (85\%)\\
7            & 215& 103& 25&  22 (88\%)\\
8            & 256& 430& 30&  24 (80\%)\\
9            & 255& 261& 40&  29 (73\%)\\
10           & 252& 382& 50&  39 (78\%)\\
11           & 255& 749& 30& 26 (85\%)\\
\bottomrule
\end{tabular}
\caption{Detailed distribution of dialogues to annotators and expert verification.}
\end{table}

\section{Additional and detailed results}
\label{additional-results}

\begin{table*}[h]
\centering
\begin{tabular}{lcccc}
\toprule
\textbf{Label} & \textbf{Precision} & \textbf{Recall} & \textbf{F1-Score} & \textbf{Support} \\
\midrule
Image       & 0.83 & 0.21 & 0.33 & 312 \\
Content        & 0.39 & 0.11 & 0.17 & 257 \\
Information      & 0.71 & 0.55 & 0.62 & 368 \\
Social norms           & 0.85 & 0.22 & 0.35 & 153 \\
Reciprocity      & 0.75 & 0.36 & 0.49 & 92  \\
Emotions          & 0.94 & 0.47 & 0.62 & 576 \\
Liking        & 0.53 & 0.53 & 0.53 & 167 \\
Authority       & 0.71 & 0.27 & 0.39 & 93  \\
Consistency    & 0.75 & 0.08 & 0.14 & 232 \\
\bottomrule
\end{tabular}
\caption{Precision, recall, and F1 scores for the Claude 3.5 Sonnet model on the SITT category detection in Polish.}
\label{tab:claude-categories}
\end{table*}

\begin{figure*}[h]
    \centering
    \includegraphics[width=1.0\linewidth]{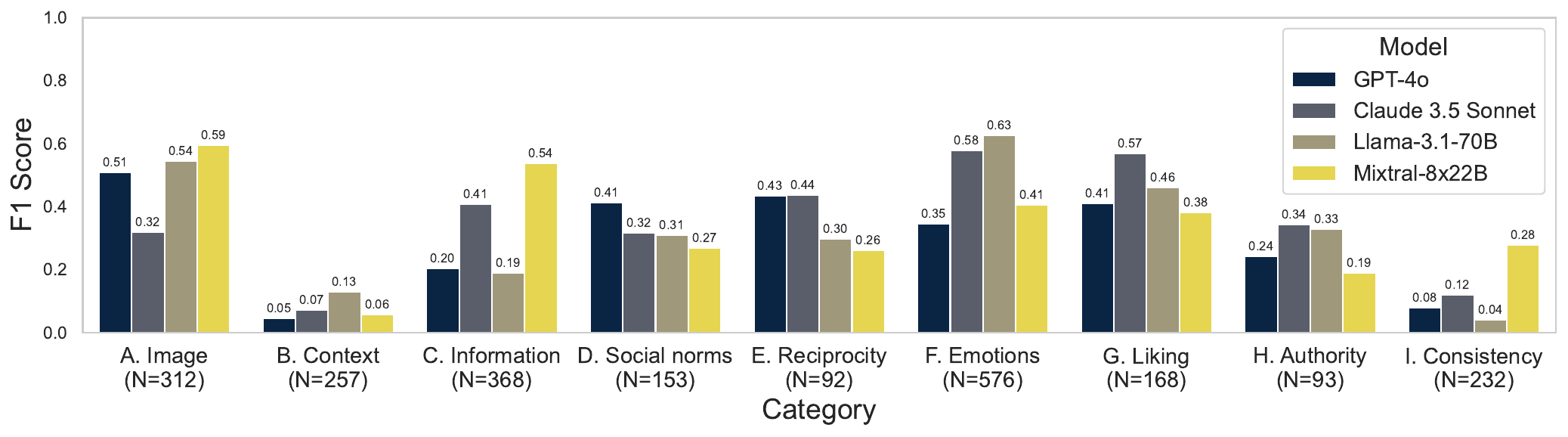}
    \caption{F1 scores of tested LLMs for classification of the SITT categories.}
    \label{fig:appendix_categories_en}
\end{figure*}

\begin{figure*}[h]
    \centering
    \includegraphics[width=1.0\linewidth]{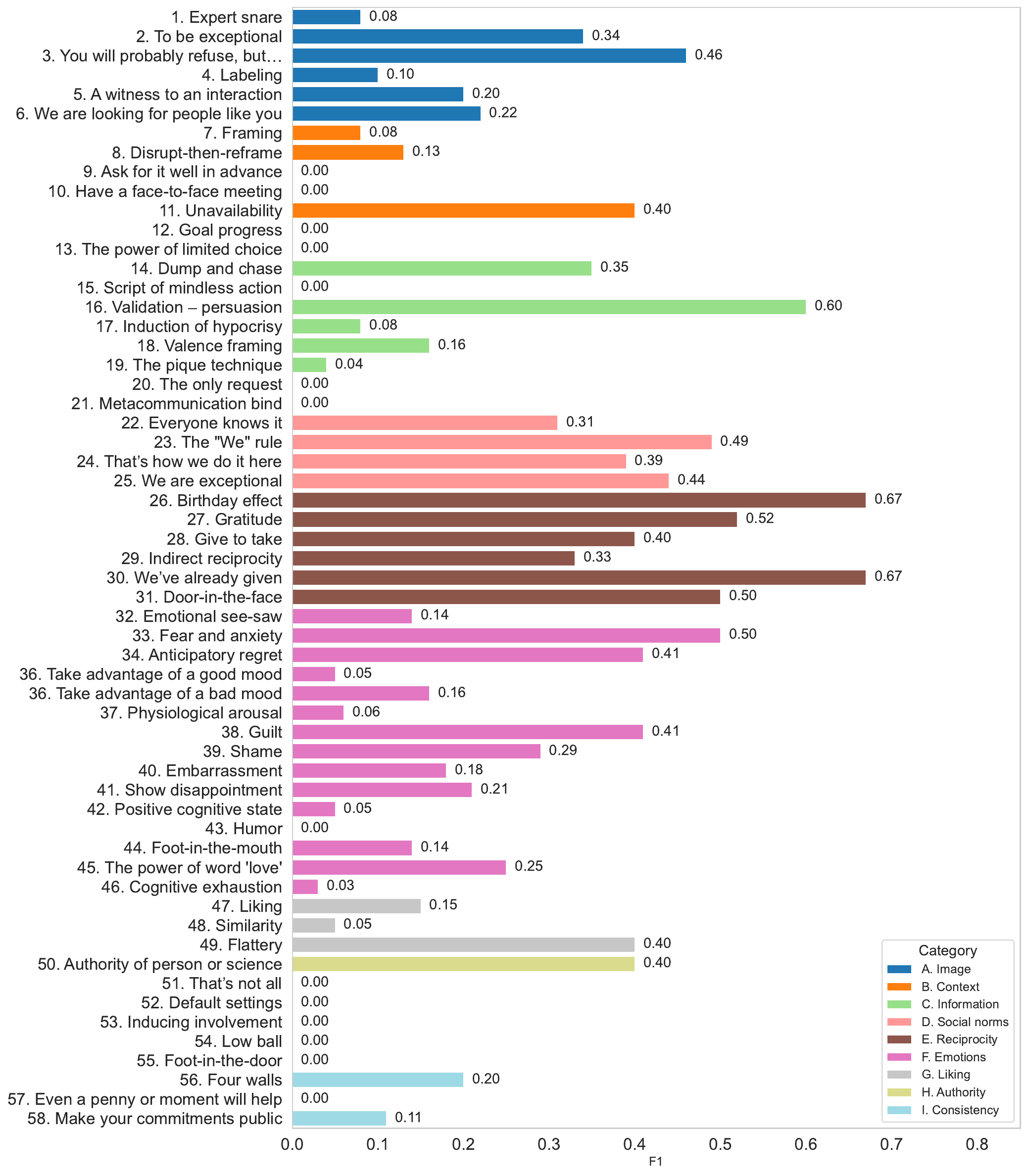}
    \caption{F1 scores of Claude Sonnet 3.5 for classification of the SITT techniques based on English dataset version.}
    \label{fig:appendix_techniques_claude_EN}
\end{figure*}

\captionsetup[longtable]{justification=centering, width=\textwidth}
\begin{longtable}{p{8.6cm}rrrr}
\toprule
\textbf{Label} & \textbf{Precision} & \textbf{Recall} & \textbf{F1-score} & \textbf{Count} \\
\midrule
\endfirsthead

\midrule
\textbf{Label} & \textbf{Precision} & \textbf{Recall} & \textbf{F1-score} & \textbf{Count} \\
\midrule
\endhead

\midrule
\multicolumn{5}{r}{\small Continued on the next page} \\
\endfoot

\endlastfoot

\multicolumn{5}{l}{\textbf{Appeal to a positive or negative image}} \\
\midrule
1. Expert snare & 0.12 & 0.17 & 0.14 & 6 \\
2. To be exceptional & 0.38 & 0.17 & 0.24 & 29 \\
3. You will probably refuse, but… & 1.00 & 0.60 & 0.75 & 5 \\
4. Labeling & 0.31 & 0.20 & 0.24 & 54 \\
5. A witness to an interaction & 0.80 & 0.67 & 0.73 & 6 \\
6. We are looking for people like you & 0.33 & 0.17 & 0.22 & 6 \\
\midrule
\multicolumn{5}{l}{\textbf{Modifying the context}} \\
\midrule
7. Framing & 0.00 & 0.00 & 0.00 & 17 \\
8. Disrupt-then-reframe & 0.00 & 0.00 & 0.00 & 9 \\
9. Ask for it well in advance & 0.00 & 0.00 & 0.00 & 4 \\
10. Have a face-to-face meeting & 0.00 & 0.00 & 0.00 & 2 \\
11. Unavailability & 0.20 & 1.00 & 0.33 & 3 \\
12. Goal progress & 0.25 & 0.17 & 0.20 & 6 \\
13. The power of limited choice & 0.08 & 0.25 & 0.12 & 4 \\
\midrule
\multicolumn{5}{l}{\textbf{Biased presentation of information and/or arguments}} \\
\midrule
14. Dump and chase & 0.26 & 0.16 & 0.20 & 43 \\
15. Script of mindless action & 0.33 & 0.01 & 0.03 & 77 \\
16. Validation – persuasion & 0.63 & 0.78 & 0.70 & 117 \\
17. Induction of hypocrisy & 0.06 & 0.40 & 0.11 & 5 \\
18. Valence framing & 0.38 & 0.13 & 0.19 & 71 \\
19. The pique technique & 0.47 & 0.08 & 0.14 & 85 \\
20. The only request & 0.00 & 0.00 & 0.00 & 16 \\
\midrule
\multicolumn{5}{l}{\textbf{Appeal to social consensus and group norms}} \\
\midrule
21. Metacommunication bind & 0.00 & 0.00 & 0.00 & 15 \\
22. Everyone knows it & 0.36 & 0.24 & 0.29 & 38 \\
23. The "We" rule & 0.56 & 0.50 & 0.53 & 18 \\
24. That’s how we do it here & 0.50 & 0.33 & 0.40 & 15 \\
25. We are exceptional & 0.89 & 0.27 & 0.41 & 30 \\
\midrule
\multicolumn{5}{l}{\textbf{Appeal to the social reciprocity}} \\
\midrule
26. Birthday effect & 0.67 & 1.00 & 0.80 & 2 \\
27. Gratitude & 0.58 & 0.78 & 0.67 & 9 \\
28. Give to take & 0.42 & 0.47 & 0.44 & 17 \\
29. Indirect reciprocity & 1.00 & 0.50 & 0.67 & 2 \\
30. We’ve already given & 0.67 & 0.80 & 0.73 & 5 \\
31. Door-in-the-face & 0.50 & 0.14 & 0.22 & 7 \\
\midrule
\multicolumn{5}{l}{\textbf{Appeal to emotions}} \\
\midrule
32. Emotional see-saw & 0.22 & 0.24 & 0.23 & 41 \\
33. Fear and anxiety & 0.81 & 0.39 & 0.52 & 235 \\
34. Anticipatory regret & 0.55 & 0.29 & 0.38 & 106 \\
36. Take advantage of a good mood & 0.14 & 0.08 & 0.11 & 24 \\
36. Take advantage of a bad mood & 0.26 & 0.16 & 0.20 & 56 \\
37. Physiological arousal & 0.00 & 0.00 & 0.00 & 19 \\
38. Guilt & 0.78 & 0.28 & 0.41 & 234 \\
39. Shame & 0.57 & 0.18 & 0.28 & 137 \\
40. Embarrassment & 0.44 & 0.10 & 0.16 & 115 \\
41. Show disappointment & 0.24 & 0.12 & 0.16 & 51 \\
42. Positive cognitive state & 0.20 & 0.05 & 0.08 & 63 \\
43. Humor & 0.33 & 0.08 & 0.12 & 13 \\
44. Foot-in-the-mouth & 0.33 & 0.25 & 0.29 & 8 \\
45. The power of word 'love'  & 0.79 & 0.27 & 0.40 & 86 \\
46. Cognitive exhaustion & 0.43 & 0.04 & 0.08 & 67 \\
\midrule
\multicolumn{5}{l}{\textbf{Appeal to sympathy, liking, connections}} \\
\midrule
47. Liking & 0.11 & 0.29 & 0.16 & 24 \\
48. Similarity & 0.03 & 0.40 & 0.06 & 5 \\
49. Flattery & 0.40 & 0.73 & 0.51 & 37 \\
\midrule
\multicolumn{5}{l}{\textbf{Appeal to authority}} \\
\midrule
50. Authority of person or science & 0.28 & 0.47 & 0.35 & 19 \\
\midrule
\multicolumn{5}{l}{\textbf{Appeal to consistency in views and/or behavior}} \\
\midrule
51. That’s not all & 0.00 & 0.00 & 0.00 & 49 \\
52. Default settings & 0.00 & 0.00 & 0.00 & 6 \\
53. Inducing involvement & 0.25 & 0.18 & 0.21 & 17 \\
54. Low ball & 0.00 & 0.00 & 0.00 & 5 \\
55. Foot-in-the-door & 1.00 & 0.40 & 0.57 & 5 \\
56. Four walls & 0.23 & 0.21 & 0.22 & 14 \\
57. Even a penny or moment will help  & 0.00 & 0.00 & 0.00 & 9 \\
58. Make your commitments public & 0.50 & 0.11 & 0.18 & 9 \\
\bottomrule
\caption{Precision, recall, and F1 scores for the Claude 3.5 Sonnet model on the SITT technique detection in Polish.}
\label{tab:calude-techniques}
\end{longtable}

\clearpage
\twocolumn
\section{Annotation guidelines}

The following section presents the instructions (in Polish) the annotators used during the annotation process.

\subsection{INSTRUKCJA ANOTACJI}

Przy ocenie poszczególnych tekstów proszę o odpowiedź na następujące
pytania:

1. Czy jest wpływ społeczny (perswazja lub manipulacja) w tekście?

2. Jaką kategorię wpływu społecznego dostrzegasz w tekście?

3. Jakie szczegółowe techniki manipulacji dostrzegasz w tekście?

\subsection{JEDNOSTKA ANOTACJI}

\textbf{I. Anotacja technik wpływu społecznego}

W procesie anotacji tekstu pod kątem występowania technik wpływu
społecznego, przyjmujemy, że podstawową jednostką anotacyjną jest
zdanie. Oznacza to, że:

\begin{enumerate}
\item
  Każde zdanie analizowane jest niezależnie pod względem potencjalnej
  obecności techniki wpływu społecznego.
\item
  Jeśli technika wyraźnie rozciąga się na więcej niż jedno zdanie (np. w
  dialogu), należy zaznaczyć kolejne zdania, które są istotne dla
  zauważenia techniki.
\item
  Jest możliwe zanotowanie jednego zdania różnymi technikami wpływu
  społecznego.
\end{enumerate}

\textbf{II. Anotacja kategorii wpływu społecznego}

W procesie anotacji przyjmujemy, że przypisanie tekstu do kategorii
wpływu społecznego następuje na poziomie całego tekstu. W jednym tekście
może wystąpić więcej niż jedna kategoria, co oznacza, że:

\begin{enumerate}
\item
  Jeśli w danym materiale jednocześnie pojawiają się różne kategorie
  wpływu społecznego to wszystkie odpowiednie kategorie powinny zostać
  przypisane.
\item
  Kategorie te nie są wzajemnie wykluczające się -- mogą
  współwystępować.
\end{enumerate}

Dzięki temu możliwa jest wielokrotna kategoryzacja jednego tekstu, co
lepiej odzwierciedla jego złożoność.

\subsection{KATEGORIE I TECHNIKI WPŁYWU
SPOŁECZNEGO}

\textbf{A. Odwoływanie się do pozytywnego/negatywnego wizerunku
odbiorcy}

\emph{Definicja:} Zespół technik wpływu społecznego, które polegają na
odwoływaniu się do samooceny odbiorcy -- jego poczucia tożsamości,
godności, moralności lub społecznego wizerunku -- w celu skłonienia go
do określonego zachowania. W przekazie wykorzystuje się zarówno
pozytywne etykietowanie (np. „jesteś odpowiedzialną osobą''), jak i
negatywne (np. „tylko ignorant by tego nie zrobił''), aby wywołać presję
do działania zgodnego z narzuconą etykietą.

\textbf{1. Technika: Usidlanie
eksperta}

\emph{Definicja:} Podkreślenie eksperckości rozmówcy podczas rozmowy,
skłania go do podtrzymywania tego wizerunku i działania zgodnie z
przypisaną rolą.

\emph{Przykład:} "Widać, że znasz się na zwierzętach, więc na pewno
zgodzisz się, że nasz produkt idealnie wpasowuje się w potrzeby kotów o
wrażliwych żołądkach."

\textbf{2. Technika: Być
wyjątkowym}

\emph{Definicja:} Podkreślenie indywidualnej wyjątkowości osoby/grupy
sprawia, że staje się bardziej skłonna do konkretnego zachowania.

\emph{Przykład:} "Ale wy jesteście nadzwyczajni. I dlatego wy (i tylko
wy) zostaniecie potraktowani w szczególny sposób"

\textbf{3. Technika: Prawdopodobnie odmówisz,
ale\ldots{}}

\emph{Definicja:} Zasugerowanie rozmówcy, że prawdopodobnie odmówi
spełnienia prośby, co paradoksalnie skłania go do jej zaakceptowania.

\emph{Przykład:} "Prawdopodobnie odmówisz, ale ciekaw jestem, czy jednak
byłbyś skłonny nam pomóc, ofiarowując datek pieniężny."

\textbf{4. Technika: Etykietowanie}

\emph{Definicja:} Przedstawianie rozmówcy jego cech w sposób, który
wywołuje w nim przekonanie o prawdziwości tej charakterystyki. W
rezultacie człowiek zachowuje się spójnie z opisaną charakterystyką.

\emph{Przykład:} "Jesteś głową rodziny. Na pewno podejmiesz decyzję,
która będzie najlepsza dla naszej rodziny."

\textbf{5. Technika: Świadek
interakcji}

\emph{Definicja:} Wykorzystywanie obecności świadka do skłonienia
rozmówcy do podjęcia decyzji -- spełnienia lub odrzucenia prośby -- w
sposób wzmacniający jego pożądany wizerunek.

\emph{Przykład:} Kasia i Tomek spacerują razem po rynku. Podchodzi do
nich wolontariusz zbierający datki na schronisko dla zwierząt. Tomek,
chcąc zrobić dobre wrażenie na Kasi, wyciąga portfel i wpłaca 50 zł.

\textbf{6. Technika: Szukamy takich jak ty}

\emph{Definicja:} Zwrócenie się z prośbą do osoby, z zaakcentowaniem, że
szuka się kogoś o konkretnych cechach, które dana osoba posiada.

\emph{Przykład:} „Szukam osób, które naprawdę dbają o środowisko, tak
jak pan,'' co zwiększa szanse na otrzymanie datku. Czy zechce pan
wesprzeć naszą akcję drobnym datkiem?

\textbf{B. Modyfikowanie
kontekstu}

\emph{Definicja:} Zespół technik wpływu społecznego polegającego na
modyfikowaniu lub wykorzystaniu elementów otoczenia, sytuacji, czasu,
miejsca lub przestrzeni w taki sposób, aby wpłynąć na postrzeganie i
decyzje odbiorców. Są one szczególnie skuteczne, ponieważ wpływają na
interpretację i emocjonalny odbiór informacji, często nieświadomie
kształtując decyzje i zachowania. Nie zmienia się informacja tylko
\textbf{kontekst}, w którym jest przedstawiana.

\textbf{7. Technika: Ramowanie}

\emph{Definicja:} Przedstawienie informacji w określonym kontekście lub
w „ramach'', które wpływają na sposób, w jaki ludzie je interpretują i
podejmują decyzje.

\emph{Przykład:} „Ten zabieg ma 90\% skuteczności'' (pozytywne
ramowanie). „Istnieje 10\% ryzyko niepowodzenia zabiegu'' (negatywne
ramowanie).

\textbf{8. Technika: Dezorientacja i zmiana ramy
interpretacyjnej}

\emph{Definicja:} Wprowadzenie osoby w stan zamieszania lub niepewności
co sprawia, że staje się mniej zdolna do racjonalnej analizy sytuacji

\emph{Przykład:} Sprzedawca przedstawia klientowi kilka różnych modeli
telefonu, za każdym razem zmieniając opinie na ich temat (np. "Ten model
jest najnowszy, ale ten z kolei ma lepszą kamerę, a ten jest bardziej
funkcjonalny"). Klient staje się zdezorientowany i trudniej mu podjąć
decyzję, a sprzedawca może wówczas łatwiej przekonać go do zakupu
jednego z modeli.

\textbf{9. Technika: Poproś z
wyprzedzeniem}

\emph{Definicja:} Proszenie o wykonanie zadania z dużym wyprzedzeniem,
ponieważ ludzie oceniają swoje przyszłe obowiązki jako mniej obciążające
niż obecne.

\emph{Przykład:} Organizator branżowej konferencji zaprasza Cię do
wygłoszenia prelekcji za osiem miesięcy. Ponieważ wydaje się to odległe,
zgadzasz się bez wahania, zakładając, że będziesz mieć więcej czasu na
przygotowanie. Gdy termin się zbliża, okazuje się, że masz napięty
harmonogram, ale już nie możesz się wycofać.

\textbf{10. Technika: Zaaranżuj
spotkanie}

\emph{Definicja:} Zachęcenie do kontaktu bezpośredniego, dzięki któremu
łatwiej budować zaufanie i zwiększyć szansę na pozytywną odpowiedź.

\emph{Przykład:} A: Dziękuję, że znalazłeś czas. Mam do Ciebie prośbę -
czy mógłbyś pomóc mi przy przygotowaniu raportu? Twoje doświadczenie
byłoby dla mnie bardzo cenne.

B: Rozumiem, chętnie pomogę.

\textbf{11. Technika
niedostępności}

\emph{Definicja:} Przypisywanie większej wartości rzeczom, które są
trudniej dostępne lub ograniczone w czasie i ilości. Wzbudzanie
poczucia, że coś jest wyjątkowe i cenne, co zwiększa pragnienie
posiadania tego.

\emph{Przykład:} Promocja tylko do 14 lipca. Spieszcie się, liczba
produktów objętych promocją ograniczona

\textbf{12. Technika: Zbliżanie się do
celu}

\emph{Definicja:} Uwydatnianie w przekazie, że realizacja celu jest
bliska, aby osoba kontynowała działanie.

\emph{Przykład:} "Spójrz przed siebie i zobacz jak jesteś blisko.
Przebędziesz jeszcze tych kilkaset metrów i jesteś na szczycie."

\textbf{13. Technika: Potęga ograniczonego
wyboru}

\emph{Definicja:} Kierowanie jednostki w pożądanym kierunku poprzez
ograniczenie liczby dostępnych opcji do wyboru.

\emph{Przykład:} "Włącz się w ratowanie naszej planety! W ramach naszej
akcji możesz: a) Zasadzić drzewo w wyznaczonym miejscu lub b) Wpłacić 20
zł na zakup sadzonki. Wybierz jeden z dwóch sposobów, w jaki możesz
pomóc przywrócić naturze to, co jej zabraliśmy. Tylko wspólnie możemy
działać skutecznie! "

\subsection{C. Tendencyjne przedstawianie informacji i/lub
argumentów}

\emph{Definicja:} Techniki w tej kategorii bazują na przedstawieniu
informacji w sposób tendencyjny, sugestywny, selektywny, dwuznaczny lub
upraszczający. W zależności od kontekstu informacja i argumenty
pochodzące od nadawcy lub odbiorcy zostają zniekształcone (np.
wzmocnione, osłabione, lub uwaga odbiorcy zostaje przekierowana). W ten
sposób obraz danej rzeczywistości zostaje zafałszowany, co może wpłynąć
na postrzeganie, opinie i decyzje odbiorcy.

\textbf{14. Technika: Zamień odrzucenie w
przeszkodę}

\emph{Definicja:} Po pojawieniu się przeszkody w realizacji prośby,
kontynuowanie dialogu przez zadawanie pytań mających na celu wyjaśnienie
przyczyn tej odmowy.

\emph{Przykład:} Rozmówca odmawia z braku czasu. Możemy zaproponować
inny termin lub krótsze spotkanie, co zwiększa szansę na akceptację.

\textbf{15. Technika: Skrypt bezrefleksyjnego
działania}

\emph{Definicja:} Dodanie jakiegokolwiek (nawet banalnego) uzasadnienia
do prośby.

\emph{Przykład:} A: Czy przesuniemy termin oddania projektu o jeden
dzień?

B: To może być problematyczne.

A: Dodatkowy dzień umożliwi zebranie wszystkich niezbędnych informacji.

\textbf{16. Technika:
Przeszkoda--perswazja}

\emph{Definicja:} Przyznanie racji rozmówcy, że jego opór przed zmianą
lub działaniem jest zrozumiały, a następnie przedstawienie argumentów
przekonujących go do podjęcia pożądanych działań.

\emph{Przykład:} Dietetyk mówi osobie odchudzającej się: „Wiem, jak
trudno zrezygnować z czekolady, bo jest pyszna i poprawia nastrój, ale
aby poprawić swoje zdrowie i uniknąć cukrzycy, warto wprowadzić zmiany w
diecie.''

\textbf{17. Technika: Indukowanie
hipokryzji}

\emph{Definicja:} Uzyskanie od osoby deklaracji popierających określone
postawy lub zachowania, a następnie wykazanie, że jej działania stoją w
sprzeczności z tymi deklaracjami.

\emph{Przykład:}

A: Wiem, że palenie jest szkodliwe, każdy to wie.

B: To czemu nadal palisz?

A: No\ldots{} to trudne do rzucenia.

B: Ale sam mówisz, że to złe dla zdrowia. Może warto spróbować jakiegoś
programu antynikotynowego?

A: Może masz rację. W sumie już o tym myślałem\ldots{}

B: Skoro już jesteśmy przy rzuceniu palenia, może zechciałbyś
zaangażować się w akcję „Czyste powietrze''?

\textbf{18. Technika: Interpretacja wyniku: zysk versus
strata}

\emph{Definicja:} Podkreślanie tego, co człowiek może stracić, jeśli
czegoś nie zrobi, jest skuteczniejsze niż mówienie o tym, co zyska,
jeśli to zrobi.

\emph{Przykład:} „Badania pokazują, że kobiety, które nie przeprowadzają
samobadania piersi, mają mniejszą szansę na wykrycie guza we wczesnej,
poddającej się leczeniu, fazie choroby"

\textbf{19. Technika: Technika wzbudzenia
zainteresowania}

\emph{Definicja:} Sformułowanie komunikatu w nietypowy sposób, aby
wzbudził zainteresowanie odbiorcy, co zwiększa prawdopodobieństwo jego
zaakceptowania.

\emph{Przykład:} 1. umówienie się na spotkanie o 16.55, zamiast na
17.00.

\textbf{20. Technika: Tylko ta jedna
prośba}
\emph{Definicja:} Podkreślenie, że prośba ma charakter jednorazowy i nie
pociąga za sobą dalszych zobowiązań.

\emph{Przykład:} "Dzień dobry, kwestuję na rzecz lokalnego hospicjum dla
dzieci, staramy się zebrać pieniądze dla jego lepszego funkcjonowania,
czy przyłączy się pan(i) do nas i wrzuci jakiś datek. To jedyna prośba,
jaką mam."

\subsection{D. Odwoływanie się do większości i/lub norm
grupowych}

\emph{Definicja:} Techniki wpływu społecznego opierające się na
odwoływaniu do norm społecznych wykorzystujące skłonność ludzi do
naśladowania zachowań większości lub przestrzegania norm akceptowanych w
danej grupie. Normy społeczne działają jako mechanizm regulujący
zachowania, a ich zastosowanie w komunikacji może skłaniać ludzi do
podejmowania działań zgodnych z oczekiwaniami grupy.

\textbf{21. Technika: Prośba o uzasadnienie
odmowy}

\emph{Definicja:} Sformułowanie prośby o wytłumaczenie się rozmówcy z
odmowy wyświadczenia nam przysługi, co jest dla niego na tyle
problematyczne, że skłania go do spełnienia naszej prośby.

\emph{Przykład}: Zwracasz się do kolegi: "Hej, potrzebuję, abyś spojrzał
na moje wyniki i dał mi jakieś wskazówki." Kolega odmawia: "Przykro mi,
ale mam za dużo pracy", Ty na to: "Rozumiem, ale czy mógłbyś mi
powiedzieć, dlaczego nie możesz mi pomóc? To dla mnie naprawdę ważne."

\textbf{22. Technika: Efekt „wszyscy to wiedzą''}

\emph{Definicja:} Odwołanie się do zdania lub zachowań większości.

\emph{Przykład:} Właściciele klubów tworzą sztuczne kolejki na zewnątrz,
aby sugerować duże zainteresowanie i wysoką jakość lokalu, co przyciąga
więcej klientów.

\textbf{23. Technika: Reguła „my''}

\emph{Definicja:} Utożsamianie się z cechami lub doświadczeniami danej
grupy w celu zwiększenia skłonności rozmówcy do spełnienia prośby,
opierając się na mechanizmie większej uległości wobec osób postrzeganych
jako członkowie własnej grupy.

\emph{Przykład:} Kolega zwraca się do współpracownika: „Wszyscy z
naszego zespołu pomagają w tym projekcie, czy możesz się przyłączyć?''
-- odwołując się do wspólnoty grupowej.

\textbf{24. Technika: U nas tak się
robi}

\emph{Definicja:} Zwrócenie uwagi na istniejącą normę społeczną
(powszechnie przyjętą zasadę postępowania) i przypomnienie jej
znaczenia.

\emph{Przykład:} Na osiedlu mieszkańcy zostają poinformowani: „W naszej
społeczności segregujemy odpady, bo tak jest u nas przyjęte,'' co
zwiększa zaangażowanie w recykling.

\textbf{25. Technika: Jesteśmy
wyjątkowi}

\emph{Definicja:} Odwołanie się do norm grupy, do której należy
odbiorca, szczególnie akcentując jej wyjątkowość. Im bardziej grupa jest
unikalna, tym silniejsza potrzeba przestrzegania jej norm, ponieważ daje
to poczucie przynależności i odróżnia jej członków od „innych''.

\emph{Przykład:} Goście hotelowi zostali poinformowani, że 75\% osób
korzystających z ich konkretnego pokoju (np. nr 215) zdecydowało się
użyć ręcznika ponownie. Odwołanie się do normy w małej, konkretnej
grupie okazało się bardziej skuteczne niż ogólne wezwania do ekologii --
aż 49,3\% gości podjęło decyzję o ponownym użyciu ręcznika, podążając za
normą „swojej'' grupy.

\subsection{E. Odwoływanie się do społecznej wzajemności}

\emph{Definicja:} Zespól technik wpływu społecznego opartych na regule
wzajemności bazują na zasadzie, zgodnie z którą ludzie czują się
zobowiązani do odwzajemnienia przysług, gestów lub korzyści, które
otrzymali od innych. Ta zasada jest głęboko zakorzeniona w normach
społecznych, ponieważ odwzajemnianie jest kluczowym mechanizmem
regulującym wymianę społeczną i budującym relacje w grupach.

\textbf{26. Technika: Efekt urodzin}

\emph{Definicja:} Kierowanie próśb do osoby, która doświadczyła w ciągu
dnia wielu przyjemnych gestów ze strony innych ludzi, wprawiających ją w
dobry nastrój.

\emph{Przykład:} Adam dostał tytuł pracownika miesiąca i otrzymuje
gratulacje przez cały dzień. Pod koniec dnia pracy koleżanka prosi go o
pomoc w realizacji jednego zadania, z którym ma problem.

\textbf{27. Technika: Okazywanie
wdzięczności}

\emph{Definicja:} Okazywanie wdzięczności osobie za wykonaną przez nią
przysługę, co nasila jej zaangażowanie w aktywność, za którą otrzymała
podziękowanie.

\emph{Przykład:} Podziękowanie za czynności związane z pracą sprawiają,
że osoba częściej o nich myśli, częściej widzi ich sens i skutki.

\textbf{28. Technika: Dać, aby wziąć}

\emph{Definicja:} Wyświadczanie drobnego gestu, przysługi drugiej
osobie, aby w przyszłości oczekiwać jego większej skłonności do
wyświadczenia nam przysługi lub spełnienia naszej prośby.

\emph{Przykład:} Zaproszenie kogoś na lunch, a za jakiś czas skierowanie
prośby o pomoc albo o zastępstwo w pracy.

\textbf{29. Technika: Zasada niebezpośredniej wzajemności}

\emph{Definicja:} Wykorzystywanie sytuacji, że dana osoba otrzymała
właśnie pomoc od kogoś innego i będzie bardziej skłonna spełnić naszą
prośbę. Osoba ta w momencie otrzymania pomocy czuje poczucie
zobowiązania, w stosunku do innej osoby niż ta, od której otrzymała
pomoc.

\emph{Przykład:} Kierowca w korku chętnie wpuszcza przed siebie
samochód, jeśli został wcześniej wpuszczony przez innego kierowcę przed
chwilą albo dużo wcześniej na innej ulicy.

\textbf{30. Technika: My już
pomogliśmy}

\emph{Definicja:} Pomaganie komuś innemu (ważnemu dla manipulowanej
osoby), by wzbudzić u niej wdzięczność i zobowiązanie do spełnienia
naszej prośby.

\emph{Przykład:} Osoba A udziela osobie B cennej rady zawodowej. Osoba B
nie ma jednak możliwości odwdzięczenia się osobie A bezpośrednio, ale
widzi, że osoba C potrzebuje pomocy w podobnej dziedzinie. Zatem osoba B
pomaga osobie C, niejako "przenosząc" gest wdzięczności od Osoby A do
Osoby B.

\textbf{31. Technika: Drzwi zatrzaśnięte przed
nosem}

\emph{Definicja:} Przedstawienie trudnej do spełnienia prośby, a po jej
odrzuceniu -- sformułowaniu drugiej, wyraźnie łatwiejszej prośby, która
jest od początku celem osoby proszącej.

\emph{Przykład:} Uczeń prosi nauczyciela o całkowite zwolnienie z
zadania domowego, wiedząc, że to niemożliwe. Po odmowie prosi o
przesunięcie terminu oddania pracy, co nauczyciel akceptuje.

\subsection{F. Odwoływanie się do
emocji}

\emph{Definicja:} To kategoria technik wpływu społecznego, która polega
na celowym wywoływaniu u odbiorcy określonych stanów emocjonalnych (np.
lęku, winy, wzruszenia, dumy, entuzjazmu), aby zwiększyć podatność na
sugestię, przekonać do określonego działania lub wpłynąć na decyzje.
Emocje te odwracają uwagę od racjonalnej analizy informacji i wzmacniają
skłonność do działania zgodnie z intencją nadawcy.

\textbf{32. Technika: Huśtawka
emocjonalna}

\emph{Definicja:} Wywołanie u rozmówcy nagłej zmiany emocji -- z
pozytywnych na negatywne lub odwrotnie; wprowadzenie go w stan
dezorientacji emocjonalnej, przez co staje się bardziej podatny na
wpływ.

\emph{Przykład:} Nauczyciel mówi uczniowi, że nie zdał ważnego egzaminu
(negatywne emocje), ale zaraz potem dodaje, że ocena została źle wpisana
i w rzeczywistości zdał (pozytywne emocje). Następnie prosi ucznia: „Czy
możesz pomóc mi uporządkować prace? To pomoże szybciej zakończyć ich
ocenianie.''

\textbf{33. Technika: Odwoływanie się do
lęku}

\emph{Definicja:} Wzbudzanie poczucia średnio nasilonego niepokoju,
strachu lub obawy.

\emph{Przykład:} „Jeśli nie wykupisz ubezpieczenia na życie, w razie
wypadku twoja rodzina zostanie bez wsparcia finansowego.''

\textbf{34. Technika: Przewidywanie
żalu}

\emph{Definicja:} wzbudzanie u rozmówcy poczucia żalu, który może
nastąpić w przyszłości z powodu wykonania lub zaniechania wykonania
działań obecnie.

\emph{Przykład:} Jeśli nie zaczniesz teraz dbać o swoje zdrowie, to za
kilka lat, gdy pojawią się problemy zdrowotne, będziesz żałować, że nic
z tym nie zrobiłeś.

\textbf{35. Technika: Wykorzystaj jego dobry
nastrój}

\emph{Definicja:} Wywołanie pozytywnego stanu emocjonalnego u odbiorcy.

\emph{Przykład:} Sprzedawca najpierw opowiada zabawną historię lub stara
się rozbawić klienta, a następnie proponuje zakup produktu,
wykorzystując jego pozytywne emocje.

\textbf{36. Technika: Wykorzystaj jego zły
nastrój}

\emph{Definicja:} Wywołanie negatywnego stanu emocjonalnego u odbiorcy.

\emph{Przykład:} Partner jest zirytowany po kłótni z kimś innym. Prosisz
go odrobną przysługę, np. wyrzucenie śmieci, mówiąc, że dzięki temu
oderwie się od swoich zmartwień.

\textbf{37. Technika: Pobudzenie
fizjologiczne}

\emph{Definicja:} Wywołanie u osoby podwyższonego pobudzenia
fizjologicznego (np. przyspieszonego bicia serca).

\emph{Przykład:} Pobudzające wyobraźnię przedstawienie zdarzenia np.
szybkiej jazdy sportowym samochodem.

\textbf{38. Technika: Poczucie winy}

\emph{Definicja:} wzbudzanie u rozmówcy poczucia winy w celu zwiększenia
skłonności rozmówcy do wyświadczenia przysługi lub spełnienia prośby
jako sposobu na obniżenie poczucia winy (jako negatywnej emocji).

\emph{Przykład:} Zostawiłeś mnie samego w tej trudnej sytuacji, a ja tak
liczyłem na twoje wsparcie i pomoc. Pomóż mi, proszę, w tym zadaniu.

\textbf{39. Technika: Poczucie
wstydu}

\emph{Definicja:} wzbudzanie u rozmówcy poczucia wstydu w celu
zwiększenia skłonności rozmówcy do wyświadczenia przysługi lub
spełnienia prośby jako sposobu na złagodzenie poczucia wstydu (jako
negatywnej emocji)

\emph{Przykład:} Twoje wyniki pracy kładą się cieniem na wizerunku
zespołu. Proszę, abyś następne zadanie zespołowe wykonał samodzielnie.

\textbf{40. Technika: Zakłopotanie}

\emph{Definicja:} Wzbudzanie u rozmówcy zakłopotania, co zwiększa jego
skłonność do spełnienia naszej prośby, aby w ten sposób mógł poczuć się
lepiej oraz poprawić swój wizerunek w oczach innych.

\emph{Przykład:} Wiem, że to może być dla Ciebie niewygodne, ale
naprawdę potrzebuję Twojej pomocy w wytypowaniu osób z naszego działu do
zwolnienia.

\textbf{41. Technika: Rozczarowanie}

\emph{Definicja:} Okazywanie rozczarowania zachowaniem rozmówcy w celu
nakłonienia go do spełnienia prośby, co może poprawić nastrój obu stron.

\textbf{Przykład:} Zawsze mogłem na Ciebie liczyć, a teraz czuję się
trochę rozczarowany, że nie masz czasu, aby mi pomóc. Czy mogę cię
prosić o wsparcie w tym zadaniu?

\textbf{42. Technika: Pozytywny stan poznawczy/ciekawość,
zaintrygowanie}

\emph{Definicja:} wzbudzenie u rozmówcy stanu zaintrygowania czy
ciekawości poprzez sztuczkę czy zagadkę, której prawdopodobnie nie
rozwiąże. W wyniku odczuwania specyficznej mieszanki ciekawości,
zaskoczenia, a jednocześnie frustracji, rozmówca jest bardziej skłonny
do spełniania próśb.

\emph{Przykład:} "Ciekaw jestem, czy uda ci się odpowiedzieć na pytanie,
które zadał mi kiedyś mój profesor". W sytuacji, gdy rozmówca nie
znajduje rozwiązania sugerujesz „Mam dla Ciebie odpowiedź. W kolejnym
kroku: Chciałbym Cię poprosić, abyś zrobił dla mnie małą rzecz''.

\textbf{43. Technika: Humor}

\emph{Definicja:} wzbudzanie uległości u jednostki poprzez 1) wplecenie
do wypowiedzi humorystycznego elementu ALBO 2) humorystyczne
formułowanie prośby. To osłabia krytyczną analizę treści komunikatu i
łatwiejszą zgodę na wyświadczenie przysługi.

\emph{Przykład:} Ej, mam wrażenie, że ta podłoga próbuje coś do mnie
powiedzieć\ldots{} ale nie rozumiem języka okruszkowego. Może byś jej
pomógł wyrazić się mopem?

\textbf{44. Technika: Stopa w ustach}

\emph{Definicja:} Wzbudzenie chęci pomocy/wyświadczenia przysługi
poprzez zobrazowanie kontrastu dobrej sytuacji odbiorcy do trudnej
sytuacji osób potrzebujących pomocy (np. bezdomni, głodujący czy
nieuleczalnie chorzy)

\emph{Przykład:} A: Jak się czujesz? B: Dziękuję, dobrze. A: Super!
Jednak nie wszyscy mają tyle szczęścia! Dzieci w Afryce głodują i
chorują na śmiertelne choroby. Możesz wesprzeć ich los.

\textbf{45. Technika: Odwoływanie się do uczucia
miłości}

\emph{Definicja:} Wywołanie u rozmówcy skojarzeń z uczuciem kochania,
miłości, silnej pozytywnej więzi

\emph{Przykład:} Prośba o datek do puszki, na której widniej napis
miłość lub love, co częściej skłonią do wrzucenia do niej pieniędzy

\textbf{46. Technika: Wyczerpanie
poznawcze}

\emph{Definicja:} Kierowanie próśb do osoby wykorzystując jej
wyczerpania fizyczne, emocjonalne lub mentalne (lub po wywołaniu
wyczerpania), co zwiększają szansę spełnienia prośby.

\emph{Przykład:} A: "Czy mógłbyś mi pomóc w czymś drobnym? To naprawdę
tylko chwila."

B: "A o co chodzi?"

A: "Super! Potrzebuję, żebyś wypełnił tę krótką ankietę, to tylko 5
pytań."

B: (z wahaniem) "Dobra, niech będzie."

\emph{(B wypełnia ankietę, zajmuje mu to więcej czasu, niż się
spodziewał.)}

A: "Dziękuję! A teraz ostatnia prośba -- czy mógłbyś dołączyć do naszej
listy uczestników? To nic wielkiego, wystarczy zaznaczyć, ile razy w
miesiącu chciałbyś pomagać przy takich projektach."

B: (zmęczony wcześniejszą aktywnością) "Uff\ldots{} dobra, wpisz mnie na
3 razy. "

\textbf{G. Odwoływanie się do sympatii, lubienia lub więzi
społecznych}

\emph{Definicja:} To kategoria technik perswazyjnych polegająca na
wykorzystywaniu emocjonalnej bliskości, sympatii lub poczucia
podobieństwa między nadawcą a odbiorcą komunikatu. Celem tych działań
jest zwiększenie skuteczności przekazu poprzez budowanie pozytywnego
nastawienia, wzbudzenie zaufania lub poczucia wspólnoty. Ludzie częściej
ulegają wpływowi osób, które lubią, z którymi się utożsamiają lub które
okazują im sympatię.

\textbf{47. Technika: Reguła
lubienia}

\emph{Definicja:} Wykorzystanie sympatii, jaką darzy nas odbiorca, do
nakłonienia go do spełnienia naszej prośby.

\emph{Przykład:} Ludzie są bardziej skłonni kupić produkt, jeśli jest on
reklamowany przez kogoś, kogo lubią.

\textbf{48. Technika: Podobieństwa}

\emph{Definicja:} Wykorzystanie cech wspólnych lub podobieństw między
osobą manipulującą a tą, do której skierowana jest prośba.

\emph{Przykład:} „Też uwielbiam grać w gry komputerowe! Widziałem, że
masz nową grę, którą chciałem wypróbować. Może pożyczysz mi tę grę?"

\textbf{49. Technika:
Komplementowanie}

\emph{Definicja:} Udzielanie rozmówcy pozytywnych, często przesadnych
komplementów celem wzbudzenia sympatii, przychylności lub wdzięczności.

\emph{Przykład:} „Naprawdę imponuje mi sposób, w jaki zarządzasz tym
projektem. Masz niesamowite zdolności organizacyjne, zawsze potrafisz
poradzić sobie w trudnych sytuacjach. Może pomożesz mi z tym zadaniem?''

\subsection{H. Odwoływanie się do autorytetu/atrybutów
autorytetu}

\emph{Definicja:} Odwoływanie się do opinii, stanowiska lub polecenia
osoby postrzeganej jako autorytet w danej dziedzinie (np. ekspert,
naukowiec, lider, lekarz), aby zwiększyć wiarygodność komunikatu i
skłonić odbiorcę do zaakceptowania określonego stanowiska lub podjęcia
działania. Również, wywoływanie posłuszeństwa lub zaufania poprzez
eksponowanie zewnętrznych oznak autorytetu, takich jak tytuł naukowy,
uniform, stanowisko, instytucja czy sposób wypowiedzi, zamiast
faktycznej wiedzy, kompetencji lub argumentów.

\textbf{50. Technika: Autorytet osoby lub
nauki}

\emph{Definicja:} Wykorzystywanie prestiżu, pozycji lub wiedzy
autorytetów w celu przekonania rozmówcy do zaakceptowania określonego
stanowiska lub argumentu.

\emph{Przykład:} Naukowcy potwierdzają, że efekt cieplarniany jest
poważnym zagrożeniem dla życia na Ziemi.

\subsection{I. Odwoływanie się do konsekwencji w poglądach i/lub
zachowaniach}

\emph{Definicja:} Zespół technik perswazyjnych, które wykorzystują
ludzką potrzebę zachowania spójności pomiędzy wcześniejszymi
deklaracjami, działaniami a aktualnymi decyzjami. Celem tych technik
jest skłonienie odbiorcy do kontynuowania wcześniej rozpoczętego
działania, wyrażonego stanowiska lub zaakceptowania bardziej
angażujących próśb, bazując na mechanizmach konsekwencji, zaangażowania
i niechęci do zmiany zdania lub kierunku działania. Techniki te
wzmacniają motywację do działania poprzez wytworzenie presji
psychologicznej wynikającej z potrzeby wewnętrznej spójności oraz chęci
bycia postrzeganym jako osoba konsekwentna i wiarygodna.

\textbf{51. Technika: To nie
wszystko}

\emph{Definicja:} Stopniowe ujawnianie elementów (korzyści)
oferty/propozycji w celu zwiększenia jej atrakcyjności dla odbiorcy

\emph{Przykład:} Nasza oferta to 100 zł za produkt, ale to nie wszystko!
Otrzymasz także darmową wysyłkę oraz dodatkowy gadżet!

\textbf{52. Technika: Reguła możliwości
zastanej}

\emph{Definicja:} Wykorzystanie ludzkiej tendencji do unikania zmiany i
pozostawania przy obecnym stanie rzeczy, szczególnie gdy podjęcie
działania wiąże się z wysiłkiem lub ryzykiem.

\emph{Przykład: Ubezpieczyciele często odnawiają polisy automatycznie, a
klienci, którzy musieliby wypowiedzieć umowę przed terminem, pozostają
przy dotychczasowym ubezpieczycielu z powodu braku aktywności.}

\textbf{53. Technika: Wzbudzanie
zaangażowania}

\emph{Definicja:} Spowodowanie, aby rozmówca zadeklarował coś
publicznie; wywołanie poczucia publicznej deklaracji.

\emph{Przykład:} W sklepie: „Czy chciałby pan tylko przymierzyć tę
kurtkę? Nie trzeba od razu kupować.'' Po przymierzeniu i uznaniu, że
kurtka ci pasuje, czujesz większą presję, by ją kupić.

\textbf{54. Technika: Niska piłka}

\emph{Definicja:} Przedstawienie rozmówcy atrakcyjnej oferty, która,
jeśli jest zaakceptowana, zostaje zmieniona na mniej korzystną.

\emph{Przykład:} Kolega prosi o krótką pomoc przy projekcie, twierdząc,
że zajmie to 5 minut. Gdy już się zgodzisz, okazuje się, że praca jest
bardziej czasochłonna, ale czujesz się zobowiązany pomóc do końca.

\textbf{55. Technika: Stopa w
drzwiach}

\emph{Definicja:} Uzyskanie zgody na spełnienie przez rozmówcę łatwej
prośby, a następnie przedstawienie mu prośby bardziej wymagającej

\emph{Przykład:} Sąsiad najpierw prosi o drobną przysługę, np.
podlewanie kwiatów podczas jego nieobecności (mała prośba). Po pewnym
czasie prosi o większą przysługę, jak np. zajęcie się jego zwierzęciem.

\textbf{56. Technika: Cztery
ściany}

\emph{Definicja:} Skłonienie rozmówcy do takich wypowiedzi, poprzez
które wpada w pułapkę konsekwencji.

\emph{Przykład: „Anita, tobie najbardziej zależy na awansie, prawda?''
-- tak, „zatem z pewnością chcesz pokazać jak bardzo jesteś kompetentna
w analizowaniu danych rynkowych'' -- tak, „z pewnością zgodzisz się na
wykonanie nowego zadania, które wymaga takich właśnie umiejętności'' --
prawdopodobnie Anita nie odmówi przyjęcia nowego zadania.}

\textbf{57. Technika: Liczy się każda poświęcona temu minuta/liczy
się każdy
grosz}

\emph{Definicja:} Nakłonienie rozmówcy do zaangażowania nawet minimalnej
ilości jakiegoś zasobu, np. czasu, pieniędzy, co spowoduje realizację
większego celu.

\emph{Przykład: „Wystarczy dosłownie złotówka -- każdy grosz ma
znaczenie i przybliża nas do celu. Nawet taka drobna kwota może pomóc
zapewnić posiłek dla osoby potrzebującej.''}

\textbf{58. Technika: Upublicznij swoje
zobowiązanie}

\emph{Definicja:} Upublicznienie swojego zobowiązania wiąże się z
większym prawdopodobieństwem, że zostanie zrealizowane.

\emph{Przykład:} Kiedy ktoś publicznie ogłasza, że zamierza ćwiczyć
codziennie przez miesiąc (np. w mediach społecznościowych), czuje
większą presję, by dotrzymać obietnicy, by nie wyjść na osobę, która nie
dotrzymuje słowa.








\end{document}